\documentclass[letterpaper,twocolumn,10pt]{article}
\PassOptionsToPackage{hyphens}{url}
\usepackage{configuration/usenix-2020-09}
\usepackage{float}
\usepackage{stfloats}
\usepackage{needspace}
\usepackage{placeins}
\usepackage{multicol}
\usepackage{booktabs}
\usepackage{array}
\usepackage{amsfonts}
\usepackage{amsmath}
\usepackage{amssymb}
\usepackage{amsthm}
\usepackage{nicefrac}
\usepackage{graphicx}
\usepackage{multirow}
\usepackage{bm}
\usepackage{enumitem}
\usepackage[font={small,bf},labelfont=bf]{caption}
\usepackage{xspace}

\let\citep\cite
\graphicspath{{resources/}{resources/figures/}{./}}
\hypersetup{
  hidelinks,
  pdftitle={CacheBridge: Efficient Cross-Model KV Transfer for Multi-Model LLM Serving},
  pdfauthor={Xingyu Qu, Siyuan Lu, Zhiyu Chen, Sheng Wang, Tao Lin},
  pdfsubject={arXiv preprint},
  pdfkeywords={cross-model KV transfer, stateful model routing, LLM serving, multi-model handoff, cache reuse}
}
\ifdefined\pdfsuppressptexinfo \pdfsuppressptexinfo=-1 \fi
\ifdefined\pdfinfoomitdate \pdfinfoomitdate=1 \fi
\ifdefined\pdftrailerid \pdftrailerid{} \fi

\newcommand{\method}{\textup{CacheBridge}\xspace}
\newcommand{\closedformmapper}{CFM\xspace}

\title{\Large\bfseries CacheBridge: Efficient Cross-Model KV Transfer for Multi-Model LLM Serving}
\author{
{\rm Xingyu Qu\textsuperscript{1,2}\thanks{Work conducted during an internship at Westlake University. Contact: \texttt{whu-quxy@whu.edu.cn}}\quad Siyuan Lu\textsuperscript{1}\quad
Zhiyu Chen\textsuperscript{3}\quad Sheng Wang\textsuperscript{2}\quad
Tao Lin\textsuperscript{1}\thanks{Corresponding author}}\\[0.35em]
{\normalsize\textsuperscript{1}Westlake University\quad
\textsuperscript{2}Wuhan University\quad\textsuperscript{3}Amazon}
}
\date{}

\begin{document}
\begin{NoHyper}    
  \maketitle
\end{NoHyper}

\begin{abstract}    
  Multi-model LLM serving uses routers, cascades and agent workflows to
  move requests across models with different costs and capabilities.
  Existing serving systems reuse KV caches only within one model, so every
  switch forces the target to re-prefill history the source already
  processed, adding time to first token (TTFT) that grows with history
  length and target size. Cross-model KV transfer can replace this
  prefill, yet prior methods are slower than re-prefill end to end. We
  present CacheBridge, the first serving system to accelerate
  heterogeneous multi-model workloads through cross-model KV transfer. It
  co-designs a compact, attention-calibrated KV mapper with systematic
  optimization for online multi-model serving, which makes model switch
  efficient on latency and memory cost.
  In Qwen3 14B$\rightarrow$32B agent serving on $\tau$-bench, it reduces
  mean TTFT by 30--98\% versus
  re-prefill at 1--24 QPS with comparable task success. Across five model
  pairs from three families, it uses 8--12$\times$ less mapper memory than
  Closed-Form Mapper and retains on average 97.94\% and 102.94\%
  of re-prefill task quality on chat/reasoning and RAG workloads,
  respectively.
  The promising results have opened new opportunities for efficient
  multi-model LLM serving through cross-model KV reuse.
\end{abstract}

%

\section{Introduction}
\label{sec:intro}

\begin{figure}[!t]
  \centering
  \includegraphics[width=\columnwidth]{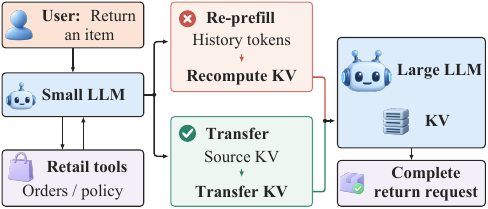}
  \caption{KV Transfer can avoid re-prefill when context switches between LLMs.}
  \label{fig:intro-scenario}
\end{figure}

Multi-model LLM serving combines models with different costs and
capabilities. Routers, cascades and agent workflows assign requests and
conversation turns to different
models~\citep{chen2024frugalgpt,ong2025routellm,wu2023autogen,zhang2026mtrouter}.
To save compute cost, they often keep routine work on a small model and
escalate only harder work to a larger one. A retail agent, for example, may
use a small model to gather user requirements and tool results before
handing the conversation, and its accumulated history, to a larger model
(Figure~\ref{fig:intro-scenario}).

This heterogeneous workload breaks the cache reuse that LLM serving relies
on for
efficiency~\citep{kwon2023pagedattention,zheng2023sglang,qin2024mooncake}.
A model cannot use another model's KV cache, so each model switch forces
the target to re-prefill history that the source has already processed. This prefill lies on the critical path of every handoff's time
to first token (TTFT), grows with history length and target size, and is
costliest under small-to-large escalation
(Section~\ref{sec:motivation-latency}). Existing serving
stacks~\citep{zheng2023sglang,qu2026haserve,li2023alpaserve,duan2024muxserve,fu2024serverlessllm,sheng2024slora,chen2024punica,wu2024dlora}
can avoid this re-prefill only by prefilling every candidate model in
advance or keeping each conversation's cache resident on every model,
which multiplies prefill computation and KV memory by the number of
models.

Cross-model KV transfer uses source KV to prepare the target's history
state. Closed-Form Mapper (\closedformmapper) uses affine maps, while
CFM+MLP uses learned nonlinear projections~\citep{heo2026crossmodel}.
Mixture-of-Translators (\textsc{MoT}) combines cache translation with
target-side history replay~\citep{lee2026mixture}, and Cache-to-Cache
(\textsc{C2C}) fuses source KV with target-computed history
KV~\citep{fu2026cache}. These methods leave two barriers to efficient
deployment. \emph{(B1) Method cost and quality.} \closedformmapper fits its
maps using all KV heads in each selected source layer, increasing mapper
storage and reconstruction work. Adding an MLP further increases the
parameter count, while fitting KV coordinates alone does not ensure task
quality. \textsc{MoT} and \textsc{C2C} do not eliminate target-side history
re-prefill. \emph{(B2) Full-path overhead.} Data
delivery and cache installation add excessive data movement,
synchronization waits and full-history temporary copies that compete with
serving for memory. In our SGLang deployment, linear \closedformmapper,
CFM+MLP and \textsc{MoT} have higher complete handoff TTFT than re-prefill at
every measured history length (Section~\ref{sec:characterization}).

To address these barriers, we present \method{}, the first serving system
to accelerate heterogeneous multi-model workloads through cross-model KV
transfer. \method{} co-designs its mapper and runtime through four
solutions.

\paragraph{(S1) Compact head assignment.}
Using linear mapping with ridge regression, \method{} reconstructs each
target KV head from one source KV head per selected source layer, reducing
mapper storage and reconstruction work compared with \closedformmapper.
It co-designs this head assignment with mapping computation placement,
enabling source-side reconstruction from local KV to reduce communication
overhead (Sections~\ref{sec:fitting-interface}
and~\ref{sec:system-placement}).

\paragraph{(S2) Attention-readout calibration.}
\method{} calibrates the value maps to match the target's attention
readouts from re-prefill, improving task quality. The corrections fold into
the existing map coefficients, preserving the compact mapper without extra
online mapping complexity (Section~\ref{sec:fitting-readout}).

\paragraph{(S3) Systematic optimization.}
\method{} pipelines reconstruction, transfer and installation with bounded
temporary memory. This systematic optimization reduces the latency and
memory cost of the handoff path
(Section~\ref{sec:system-concurrency}).
Together, these optimizations provide the foundation for efficient online
serving.

\paragraph{(S4) Online serving policy.}
Based on this system design, \method{} proposes a workload- and
resource-aware online serving policy to coordinate KV transfer with ongoing
inference in multi-model serving. It selects the faster handoff path for
each request and adapts transfer concurrency and memory allocation to
serving load
(Section~\ref{sec:system-policy}).

\textbf{For task quality,} \method{} achieves the best overall task quality
and the most robust retention among the transfer baselines across five
model pairs from three families, retaining on average 97.94\% and 102.94\%
of re-prefill quality on chat/reasoning and RAG workloads, respectively.
\textbf{For system efficiency,} \method{} reduces mean handoff TTFT to about
one-sixth of re-prefill for 32K histories on Qwen3 14B$\rightarrow$32B.
On the $\tau$-bench~\citep{yao2024taubench} retail-agent workload under
Poisson request arrivals, \method{} reduces mean TTFT by 30--98\% at
1--24 QPS with task success comparable to re-prefill. Across pairs, it uses
8--12$\times$ less mapper memory than \closedformmapper
(Section~\ref{sec:evaluation}).

The paper makes four contributions:
\begin{itemize}[leftmargin=*,nosep]
    \item We first profile the online serving cost of KV transfer methods
    in heterogeneous multi-model workloads and show that existing KV
    transfer methods do not yield an efficient online handoff
    (Section~\ref{sec:characterization}).
    \item We design a compact, attention-calibrated KV mapper that combines
    efficient KV reconstruction with high task quality
    (Section~\ref{sec:fitting}).
    \item We build \method{}, the first serving system to accelerate
    heterogeneous multi-model workloads through cross-model KV transfer
    (Section~\ref{sec:implementation}).
    \item \method{} achieves the best overall task quality among the
    baselines across diverse datasets and model pairs, and substantially
    reduces latency and improves throughput on the agent serving workload
    (Section~\ref{sec:evaluation}).
\end{itemize}

\section{Background and Preliminaries}
\label{sec:background}

\subsection{Multi-model Handoffs}
\label{sec:background-workloads}

KV caches are model-specific: each layer stores key and value projections
of the processed history. Differences in layer counts, KV heads, head
dimensions and learned representations prevent direct reuse across models.
Routers, cascades and agents switch models over histories containing
prompts, conversation turns, retrieved passages and tool
results~\citep{ong2025routellm,zhang2026mtrouter,chen2024frugalgpt,wu2023autogen}.
The router selects the target, and the serving system prepares its history
cache. Re-prefill repeats history computation on the target at each switch,
while concurrent handoffs compete with inference for compute and memory.
This paper considers models within the same family that share a tokenizer,
so history positions index the same tokens in both models.

\subsection{Representative KV Transfer Methods}
\label{sec:background-methods}

\paragraph{Training-free CFM.}
Closed-Form Mapper (\closedformmapper) fits separate affine maps for each
target layer's key and value heads using closed-form ridge
regression~\citep{heo2026crossmodel}. It selects predictive source layers
for each target layer and concatenates all KV heads in those layers as
input. We refer to this assignment as \emph{Full-Head}, and to the number
of source heads read per target head as \emph{mapper width}. Source rotary
position embeddings (RoPE)~\citep{su2021roformer} are removed before key mapping and target
rotations restored afterward. Fitting uses paired source and target KV
from calibration text while both models remain frozen. At handoff, the
maps directly reconstruct the target's full history cache. The trained
CFM+MLP variant uses nonlinear projections for the same reconstruction.
Training-free refers to the closed-form fit, which still requires paired
calibration caches. The same coefficients are reused at every history
position, so mapper storage depends on the selected layers and head
dimensions, while reconstruction work grows with history length.

\paragraph{Training-based MoT.}
Mixture-of-Translators (\textsc{MoT}) trains multiple cache translators with
token-level routing and a Context Correction Loss that aligns replayed
target computation with native target computation~\citep{lee2026mixture}.
Routing combines translators separately for each token. The correction
loss supervises replayed target states, so translator training accounts for
the target computation following cache translation.
Translated KV is injected into selected target layers, while
\emph{target-side history replay} processes the history to construct the
remaining layers' caches. To reduce replay cost, \emph{source guidance}
selects history positions using source attention scores, and target replay
attends to those positions with sparse attention. Thus replay retains
target-side history computation after mapping. Cache-to-Cache
(\textsc{C2C}) instead fuses source KV with target-computed history KV,
also retaining target-history computation~\citep{fu2026cache}.

KV coordinate error, measured by normalized mean squared error (NMSE),
compares reconstructed keys and values with re-prefill. The target uses
them through \emph{attention readouts}: query-dependent weighted sums of
cached values, with weights computed from queries and cached keys.
Deployment also requires data delivery, tensor-parallel (TP) communication
among GPU ranks, and cache installation in the target engine's KV-head
layout before its scheduler can execute the query. These steps can involve
host-side receive and acknowledgment waits and history-length temporary
buffers. Section~\ref{sec:characterization} profiles mapping quality and
this complete handoff path.

\section{Motivation and Deployment Challenges}
\label{sec:characterization}

We profile the online serving cost of existing KV transfer methods in
the multi-model workloads described in Section~\ref{sec:background}. We
compare linear
\closedformmapper, \closedformmapper with learned MLP projections
(CFM+MLP)~\citep{heo2026crossmodel} and \textsc{MoT}~\citep{lee2026mixture}
against re-prefill on Qwen3 14B$\rightarrow$32B, with each model on four
H800 GPUs under four-way tensor parallelism (TP4).

\subsection{Handoff Latency}
\label{sec:motivation-latency}

All comparisons below use end-to-end handoff TTFT measured with both
models deployed in SGLang engines.
Prior evaluations of these methods do not cover this complete serving path,
leaving the deployment bottlenecks and achievable latency gains of
cross-model KV transfer uncharacterized.
Handoff TTFT runs from a committed model switch, with the source cache and
query available, to the first target token returned to the host. For
transfer, it includes data delivery, reconstruction, installation and any
target-side history replay before query processing.

\begin{figure}[!htbp]
    \centering
    \includegraphics[width=\columnwidth,trim=0 5bp 0 1bp,clip]{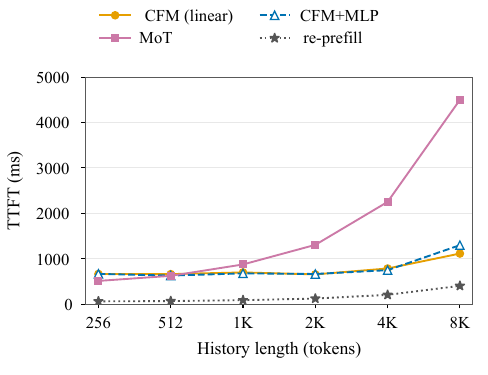}
    \caption{All three transfer methods have higher median handoff TTFT than
    re-prefill at every length.}
    \label{fig:e1-handoff-latency}
\end{figure}

Re-prefill TTFT grows with history length, from about 60\,ms at 256 tokens
to about 400\,ms at 8K tokens (Figure~\ref{fig:e1-handoff-latency}). All
three transfer methods are slower than re-prefill at every measured length;
at 8K tokens, linear \closedformmapper, CFM+MLP and \textsc{MoT} take about
2.8$\times$, 3.2$\times$ and 11$\times$ the TTFT of re-prefill. In this SGLang deployment, the complete paths of
these methods do not turn cross-model reuse into an online latency gain.

\begin{figure}[H]
    \centering
    \includegraphics[width=\columnwidth]{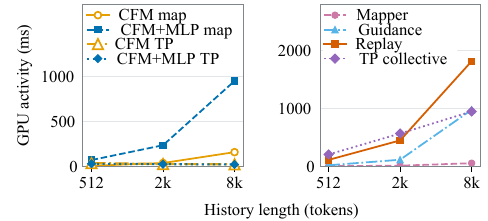}
    \caption{Mapping dominates CFM at long histories (left). MoT stages
    (right) are KV mapping, source guidance, target-history replay and
    tensor-parallel communication.}
    \label{fig:e1-stage-profile}
\end{figure}

In the GPU time profile (Figure~\ref{fig:e1-stage-profile}), mapping grows
with history length and dominates both \closedformmapper variants at long
histories, and
the MLP raises mapping time about sixfold at 8K tokens. \textsc{MoT} spends
little time in its mapper; source guidance, target-side replay and TP
collectives dominate instead, and replay grows from about
0.1\,s at 512 tokens to 1.8\,s at 8K, so \textsc{MoT} reintroduces the
target-side history computation that transfer is meant to remove.

\paragraph{Observation 1.} \emph{Transfer pays off only when its complete
path costs less than re-prefill, and existing methods spend that budget on
mapping, delivery and retained history computation.} This motivates compact
mapping (S1), systematic optimization (S3) and online path selection (S4).

\subsection{Mapper Width and Data Movement}
\label{sec:motivation-support}

Since mapping dominates \closedformmapper's GPU time at long histories, we
examine whether fewer source heads can preserve task quality.
The number of source heads that each target head reads sets the mapper's
width. We fit \closedformmapper maps with one, two, four and
eight source heads per target head and evaluate them on 512 HellaSwag
examples. For each
width below eight, we enumerate source-head combinations and keep the best.
The best single head is the same-index source head, and eight heads use all
source heads (Full-Head), as in \closedformmapper.

Accuracy rises from 75.00\% with one head to 79.69\% with the best two-head
combination, then falls to 78.13\% with four or eight heads,
although K and V NMSE are lowest
with four or eight heads. Lower coordinate error does not yield higher task
accuracy.

\begin{figure}[!htbp]
    \centering
    \includegraphics[width=\columnwidth,trim=0 2.25bp 0 8.25bp,clip]{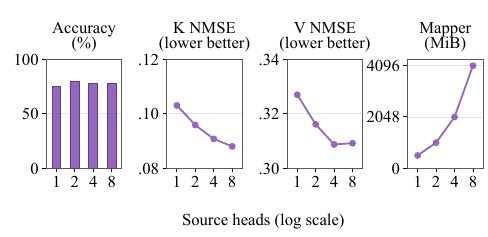}
    \caption{Before calibration, HellaSwag accuracy peaks at two source heads per target head;
    storage grows linearly.}
    \label{fig:head-support-hellaswag}
\end{figure}

\begin{figure}[!htbp]
    \centering
    \includegraphics[width=\columnwidth,trim=0 5.75bp 0 2.5bp,clip]{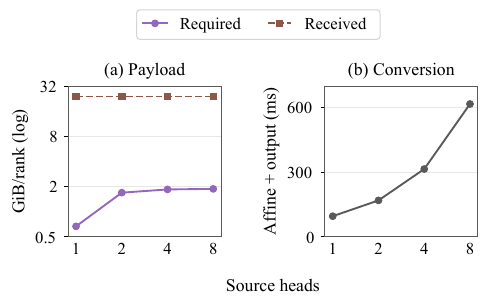}
    \caption{At 8K tokens, received payload does not shrink with mapper
    width.}
    \label{fig:e3-support-delivery}
\end{figure}

Costs grow steadily with width. Mapper storage rises eightfold
from 512\,MiB to 4,096\,MiB (Figure~\ref{fig:head-support-hellaswag}).
At 8K tokens, the source state that a target
rank's maps read, one copy per distinct source layer and KV head with FP32
keys and BF16 values, grows from 0.67 to 1.87\,GiB
(Figure~\ref{fig:e3-support-delivery}(a)), and conversion time, measured
separately over the affine maps and output conversion, grows about sixfold
(Figure~\ref{fig:e3-support-delivery}(b)). The measured path, however,
receives about 24\,GiB per rank at every width, 13--36$\times$ what the maps
read, so a narrower mapper does not reduce actual data movement.

\paragraph{Observation 2.} \emph{Increasing mapper width increases storage,
required source-state payload and reconstruction work without consistent
quality gains.}
This motivates compact head assignment (S1), co-designed with source-side
mapping (Sections~\ref{sec:fitting-interface} and~\ref{sec:system-placement}).
With attention-readout calibration (S2, Section~\ref{sec:fitting-readout}),
the one-head mapper retains 104.00\% of re-prefill quality on HellaSwag for
this pair, versus 98.75\% for Full-Head \closedformmapper
(Table~\ref{tab:accuracy-complete-main} in
Appendix~\ref{sec:appendix-complete-accuracy}).

\subsection{Reconstructed-State Quality}
\label{sec:motivation-quality}

We compare re-prefill and Full-Head \closedformmapper on 64 examples each
from MMLU-Redux and HotpotQA-E, whose histories span 128--429 and
8,923--16,371 tokens, respectively.

\begin{table}[!htbp]
    \centering
    \small
    \setlength{\tabcolsep}{4pt}
    \caption{Task scores and Full-Head CFM readout NMSE.}
    \label{tab:long-context-cfm-diagnostic}
    \begin{tabular}{@{}lrrr@{}}
        \toprule
        Dataset & re-prefill score & CFM score & NMSE \\
        \midrule
        MMLU-Redux & 56.25 & 79.69 & 0.3287 \\
        HotpotQA-E & 58.37 & 35.37 & 0.4259 \\
        \bottomrule
    \end{tabular}
\end{table}

\begin{figure*}[!t]
  \centering
  \includegraphics[width=\textwidth]{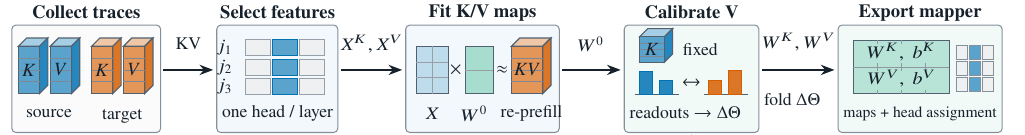}
  \caption{Offline fitting: one source KV head per selected layer, then
  readout calibration of the value maps.}
  \label{fig:fitting-overview}
\end{figure*}

Full-Head \closedformmapper exceeds re-prefill on MMLU-Redux but loses
23 points on HotpotQA-E,
retaining 60.6\% of the re-prefill score although it reads every source head
(Table~\ref{tab:long-context-cfm-diagnostic}). The target consumes the
reconstructed K/V only through query-dependent attention, so we compare its
attention readout $O=AV$, where $A$ holds the query's attention weights over
cached keys and $V$ the cached values, under reconstruction and under
re-prefill (Figure~\ref{fig:long-context-cfm-readout} in
Appendix~\ref{sec:appendix-fitting-diagnostics}). Mean readout NMSE is
higher where quality drops: 0.4259 on HotpotQA-E versus 0.3287 on
MMLU-Redux.

\paragraph{Observation 3.} \emph{Full-Head mapping can incur substantial
quality loss on long-context QA.} Our broader evaluation examines this gap
across tasks and model pairs (Section~\ref{sec:eval-accuracy}). The higher
readout error on HotpotQA-E motivates attention-readout calibration
(S2, Section~\ref{sec:fitting-readout}), whose quality contribution is
evaluated in Section~\ref{sec:eval-ablation}.

\subsection{Delivery and Installation}
\label{sec:motivation-delivery}

Beyond moving 13--36$\times$ the source state that the maps read
(Section~\ref{sec:motivation-support}), the
handoff path waits on delivery and holds history-length buffers during
installation. In a profiled
8K-token run, \closedformmapper spends about 69\% of its TTFT in
host-side receive and acknowledgment spans that wait for peer ranks, more
than its mapping and TP activity combined. Its history-length buffers
compete with serving for the target's free memory and cap how many
transfers can run concurrently.

\paragraph{Observation 4.} \emph{The existing handoff path moves far more
source state than the maps read, waits on peer ranks and holds
history-length buffers.} These costs motivate source-side mapping (S1),
pipelined transfer (S3) and resource-aware admission
(S4, Section~\ref{sec:implementation}).

Sections~\ref{sec:fitting} and~\ref{sec:implementation} develop S1--S4 by
co-designing mapper quality with handoff latency and memory.

\section{Fitting a Deployable KV Mapper}
\label{sec:fitting}
\label{sec:design}

\method{} addresses Observations~1--3 through compact head assignment
(S1, Section~\ref{sec:fitting-interface}) and attention-readout calibration
(S2, Section~\ref{sec:fitting-readout})
(Figure~\ref{fig:fitting-overview}). Both LLMs remain frozen. The head
assignment is co-designed with source-side mapping computation placement
(Section~\ref{sec:system-placement}).

\subsection{Compact Head Assignment and Initial Fit}
\label{sec:fitting-interface}

To realize S1 within Observation~1's handoff latency budget
(Section~\ref{sec:motivation-latency}), \method{} reduces mapping work with
the narrow head assignment motivated by Observation~2
(Section~\ref{sec:motivation-support}): each target KV head reads one source
KV head per selected layer.

\paragraph{Head assignment.}
We retain \closedformmapper's source-layer selection and RoPE-aware affine
maps (Section~\ref{sec:background-methods}). Let $\mathcal S_\ell$ be the
source layers selected for target layer $\ell$, and let $a(h)$ assign a
source KV head to target head $h$. In grouped-query attention (GQA)~\citep{ainslie2023gqa}, $a(h)$
follows the fixed ownership of query heads by KV heads.
All five main pairs have equal KV-head counts and use $a(h)=h$.
Other group-count relationships use a fixed map derived from architecture
metadata (Appendix~\ref{sec:appendix-architecture-extension}). The main pairs
reconstruct the full history cache directly, with local-attention and
recurrent extensions in the appendix.
For $Z\in\{K,V\}$ and history position $t$,
\begin{equation}
  \begin{split}
    x^Z_{\ell,h,t}
      &=\operatorname{concat}_{j\in\mathcal S_\ell} Z^S_{j,a(h),t},\\
    \widehat Z^T_{\ell,h,t}
      &=x^Z_{\ell,h,t}W^Z_{\ell,h}+b^Z_{\ell,h}.
  \end{split}
  \label{eq:fitting-affine}
\end{equation}
Here $S$ and $T$ denote the source and target, respectively.
Hats denote reconstruction, and $W^Z,b^Z$
are the key- or value-map weights and bias. Keys are mapped
in content coordinates: source positional rotations are removed, and target
rotations are restored at the target history positions.

For $k=|\mathcal S_\ell|$ selected layers and source head width $d_s$, the
feature width falls from $kH_sd_s$ with all source heads to $kd_s$, where
$H_s$ is the number of source KV heads. At the same layer selection, the
non-bias coefficient count and the affine multiply--add count fall by a
factor of $H_s$; for Qwen3 14B$\rightarrow$32B ($H_s=8$), mapper storage
falls from 4,096\,MiB to 512\,MiB. This shared head assignment also places
each target shard's inputs on its matching source rank
(Section~\ref{sec:system-placement}).

\begin{figure}[!b]
  \centering
  \includegraphics[width=\columnwidth,trim=0 3.25bp 0 0.5bp,clip]{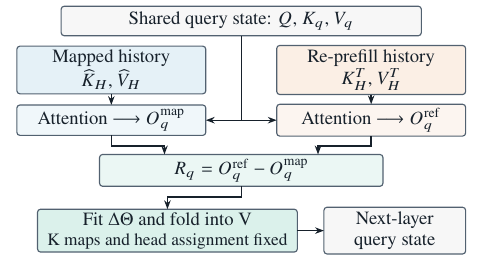}
  \caption{Value-map calibration at one target layer.}
  \label{fig:readout-calibration}
\end{figure}

\paragraph{Source layers and initial fit.}
We run both models on the same 500 FineWeb-Edu~\citep{penedo2024fineweb}
calibration sequences and collect KV vectors at matching token positions.
For each target layer $\ell$, we fit centered linear regressions from each
source layer's KV vectors to the target's, separately for K, V and each
matched KV head. We average the resulting $R^2$ scores over K, V and the
matched heads, then select the $k$ highest-scoring source layers as
$\mathcal S_\ell$. All heads share this set within a target layer, while
different target layers select their own sets. We choose one $k$ per model
pair on independent validation data to balance task quality against mapper
storage and reconstruction work (Appendix~\ref{sec:appendix-calibration}).

For each target KV head, we concatenate its assigned source-head vectors
from the selected layers and fit separate affine ridge maps to the
target's re-prefill K and V, following
\closedformmapper~\citep{heo2026crossmodel}. These maps initialize the
attention-readout calibration. Appendix~\ref{sec:appendix-calibration}
details the paired data, layer scores, ridge solver and sensitivity analysis.

\subsection{Calibrating Value Maps with Attention Readouts}
\label{sec:fitting-long-context}
\label{sec:fitting-readout}

To realize S2 in response to Observation~3
(Section~\ref{sec:motivation-quality} and
Appendix~\ref{sec:appendix-fitting-diagnostics}),
\method{} calibrates the value maps against the target's attention readouts,
keeping the key maps and head assignment fixed.

\paragraph{Calibration handoffs.}
We form document--query pairs $(H,q)$ from 24 QuALITY documents,
with four queries per document. As in an online handoff, the source
processes $H$, the initial mapper reconstructs target history KV, and the
target processes the appended query $q$. Offline target re-prefill
provides the reference history KV.
Appendix~\ref{sec:appendix-calibration} details the calibration setup
and evaluation exclusions.

\paragraph{Local readout target.}
Figure~\ref{fig:readout-calibration} shows the comparison at one target
layer. Let $Q$ denote the current query state and $(K_q,V_q)$ the fresh KV
computed from it. We compare
\begin{equation}
  \begin{split}
    O^{\mathrm{map}}_q
      &=\operatorname{Attn}(Q,[\widehat K_H,K_q],
                              [\widehat V_H,V_q]),\\
    O^{\mathrm{ref}}_q
      &=\operatorname{Attn}(Q,[K^T_H,K_q],[V^T_H,V_q]),\\
    R_q&=O^{\mathrm{ref}}_q-O^{\mathrm{map}}_q.
  \end{split}
  \label{eq:fitting-readout-target}
\end{equation}
The reference substitutes target history KV from re-prefill at the same
query state, and readouts are compared before the output projection.

\paragraph{Value-map correction.}
Let $A^{\mathrm{map}}_{q,H}$ contain the mapped branch's attention weights on
history positions, normalized over all causally visible history and query
positions. Let $\bar X^V_H=[X^V_H,\mathbf 1]$ augment the selected source V
features with an intercept. Because the key maps are fixed,
$A^{\mathrm{map}}_{q,H}$ does not depend on the correction, and an update
$\Delta\Theta$ to the value map's weights and bias changes the readout
linearly:
\begin{equation}
  \Delta O_q=A^{\mathrm{map}}_{q,H}\bar X^V_H\Delta\Theta.
  \label{eq:fitting-readout-change}
\end{equation}
With readout features $D_q=A^{\mathrm{map}}_{q,H}\bar X^V_H$, we fit
one shared correction per target layer and KV head over the calibration
queries $\mathcal Q$:
\begin{equation}
  \begin{aligned}
    \min_{\Delta\Theta}\quad&
      \sum_{q\in\mathcal Q}w_q
      \|D_q\Delta\Theta-R_q\|_F^2\\
      &\quad+\lambda_{\mathrm{ro}}\|\Delta\Theta\|_F^2\\
    \text{subject to}\quad&
      \operatorname{tr}(\Delta\Theta^{\mathsf T}M\Delta\Theta)
      \leq 0.01E_0.
  \end{aligned}
  \label{eq:fitting-readout}
\end{equation}
Here $w_q$ weights documents equally and $\lambda_{\mathrm{ro}}$
regularizes the correction. With source-feature second moment $M$, the
constraint limits the update's energy on the initial fitting text to
1\% of the initial value residual energy $E_0$.
Appendix~\ref{sec:appendix-calibration} details the row construction,
query weighting, energy metric and constrained solver.

\begin{figure*}[!tbp]
  \centering
  \includegraphics[width=\textwidth,trim=0 1.75bp 0 2.25bp,clip]{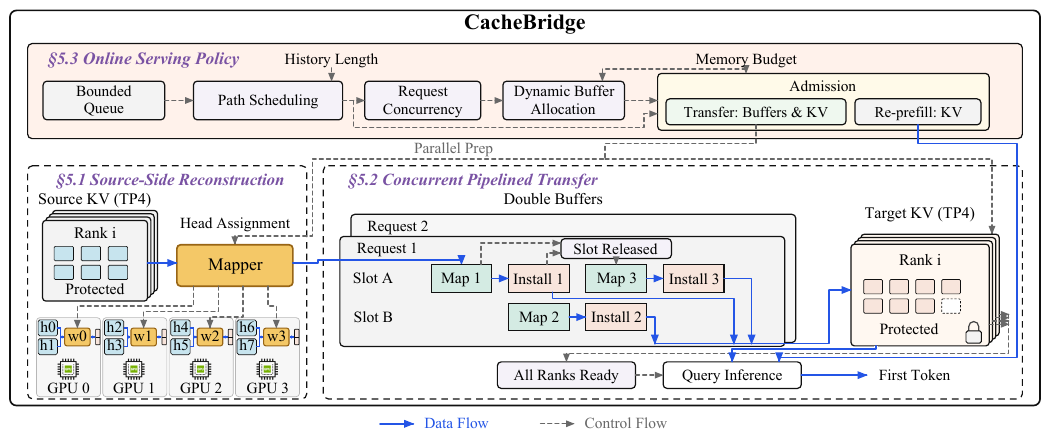}
  \caption{\method{}: System Design and Online Serving}
  \label{fig:system-overview}
\end{figure*}

\paragraph{Layer-ordered calibration.}
At each target layer, we solve one correction per KV head shared across the
calibration queries, fold it into the value map, and propagate each query
state through the corrected layer. The next layer is calibrated at these
states, so each local fit accounts for the corrections already applied
along the target's query computation.

We export the final key and value maps, with all readout corrections folded
in, together with the selected source layers and head assignment.

\section{\method{} System Design}
\label{sec:system}
\label{sec:implementation}

The online runtime completes S1 through source-side reconstruction
(Section~\ref{sec:system-placement}), realizes S3 through concurrent
pipelined transfer (Section~\ref{sec:system-concurrency}), and implements
S4 through a workload- and resource-aware serving policy
(Section~\ref{sec:system-policy}). Each model runs in its own
tensor-parallel engine, coordinated through a shared pool of transfer
buffer sets.

\subsection{Source-Side Reconstruction}
\label{sec:system-placement}

Observation~4 shows that the existing handoff path moves 13--36$\times$ the
source state that the maps read and waits on peer ranks
(Section~\ref{sec:motivation-delivery}). To complete S1, \method{} places
mapping computation on source GPUs and reconstructs target KV from resident
history. Only target
KV, which the target must hold anyway, crosses between models, and the
target's compute stays available for serving. Reconstruction on the target
would instead deliver the mapper's source inputs and place reconstruction
compute and temporary memory on the target
(Figure~\ref{fig:system-overview}).

\paragraph{GPU locality from head assignment.}
The head assignment (Section~\ref{sec:fitting-interface}) is co-designed
with mapping computation placement and tensor parallelism: each target KV head reads only its assigned source
head in each selected source layer, so with aligned TP head partitions every
input of a target shard resides on the matching source GPU. For Qwen3
14B$\rightarrow$32B under TP4, source rank $r$ holds source KV heads $2r$ and
$2r{+}1$ and reconstructs target heads $2r$ and $2r{+}1$, the complete shard
of target rank $r$, without reading KV from any other GPU. A full-head mapper
(like \closedformmapper~\citep{heo2026crossmodel}) would instead make
every target head depend on all source heads of its
selected layers, so each source rank would first gather three quarters of
the selected source KV from its peers.

\paragraph{Pairwise delivery.}
Because source rank $r$ produces exactly the KV that target rank $r$ stores,
each target rank copies blocks only from its matching source rank. The rank
pairs transfer in parallel, and each block's ready and acknowledgment
signals stay within its pair, so no block waits on peer ranks. Each target
rank receives only its own shard, 0.5\,GiB at 8K tokens for Qwen3
14B$\rightarrow$32B, about 48$\times$ less than the 24\,GiB per rank on the
existing path (Section~\ref{sec:motivation-support}).

\subsection{Concurrent Pipelined Transfer}
\label{sec:system-concurrency}

Observation~4 also shows that the existing path holds history-length
buffers during installation. To realize S3, \method{} installs
reconstructed KV block by block and pipelines reconstruction, transfer and
installation to reduce handoff latency and bound temporary memory. The
resulting path supports concurrent pipelines admitted by S4
(Section~\ref{sec:system-policy}).

\paragraph{Direct installation.}
The source reconstructs 2,048-token blocks to balance TTFT and buffer
memory (Table~\ref{tab:system-ablation-block-size} in
Appendix~\ref{sec:appendix-system-ablation}), installing each directly into
the target's allocated paged KV storage. Avoiding a full-history staging
copy bounds temporary memory by block size and admitted transfers.

\paragraph{Pipeline overlap.}
Each transfer alternates between output slots A and B. While the target
reads a block from one slot, the source reconstructs the next block into
the other, so reconstruction, transfer and installation of consecutive
blocks overlap.

\paragraph{Completion and buffer reuse.}
A slot is released once source reconstruction and target reads of its block
both finish. Once every target rank confirms installation of its complete
history shard, the target publishes the pages and runs query inference.
Source history remains protected throughout reconstruction, and target KV
during query processing and decoding.

\subsection{Online Serving Policy}
\label{sec:system-policy}

To realize S4, the policy combines history length for path selection,
available GPU memory for admission, and serving demand for transfer
concurrency and buffer allocation.

\paragraph{Path scheduling.}
Re-prefill is faster for short histories and transfer for long ones
(Section~\ref{sec:eval-system}). The crossover history length is profiled
offline and lies near 2K tokens for Qwen3 14B$\rightarrow$32B; the policy
routes shorter histories to re-prefill and considers longer ones for
transfer.

\paragraph{Request concurrency.}
Each admitted transfer runs its pipeline in its own buffer set, so
concurrent transfers progress independently. A set holds an input buffer and
two output slots for 2,048-token blocks, so its fixed size turns buffer
memory into an explicit concurrency limit. With per-request footprint
$m_{\mathrm{buf}}$ and memory $M_{\mathrm{map}}$ left after ordinary serving
reservations, both per source TP rank, the pool supports up to
$K_{\max}=\lfloor M_{\mathrm{map}}/m_{\mathrm{buf}}\rfloor$ concurrent
transfers on the smallest rank, and target KV capacity further bounds
admission. The coordinator assigns a free set to each long-history request;
if none is available, the request falls back to re-prefill.

\paragraph{Dynamic buffer allocation.}
Completed transfers return their sets to the pool for reuse. The coordinator
grows the pool within the current budget and reclaims unused sets when
serving needs memory or after an idle interval, once outstanding mapper uses
and buffer leases finish on all ranks.

\subsection{Implementation}
\label{sec:system-implementation}

We implement \method{} on SGLang v0.4.10.post2 with independent engines on
disjoint GPU groups. About 50 modified scheduler lines support source-prefix
protection, target-page allocation and prefix publication.
Native attention and decoding are unchanged. Reconstruction uses PyTorch
CUDA for page gathering, RoPE removal/restoration, FP32
affine maps and BF16 output, without a custom fused kernel. Intra-server
installation uses \texttt{cudaMemcpyPeerAsync}. A Python coordinator runs
the policy with one or two buffer sets.
Appendix~\ref{sec:appendix-implementation} details intra-server coordination.

\paragraph{Cross-node execution.}
For cross-node execution, we deploy the source and target engines on
separate servers, each with eight NVIDIA H800 80-GB GPUs running TP8.
Each server uses four active ConnectX-7 InfiniBand ports at 400\,Gb/s per
port, providing a nominal aggregate link capacity of 1.6\,Tb/s.
Eight corresponding source--target GPU-rank pairs communicate through NCCL
point-to-point operations with GPUDirect RDMA, replacing the intra-server
CUDA IPC and \texttt{cudaMemcpyPeerAsync} transport.
Reconstructed BF16 KV blocks travel from source GPU memory through the
source NIC, the InfiniBand fabric, and the target NIC directly into
allocated target GPU KV-pool pages, without host-memory staging of KV
payloads.

\section{Evaluation}
\label{sec:evaluation}

We evaluate task quality (Section~\ref{sec:eval-accuracy}) and serving
performance (Section~\ref{sec:eval-system}). Ablations
isolate compact mapping, attention-readout calibration, handoff-path
optimization and the online policy (Section~\ref{sec:eval-ablation}).

\subsection{Experimental Setup}
\label{sec:eval-setup}

\paragraph{Hardware and software.}
Single-server experiments use one 8-GPU NVIDIA H800 server, with source and target on
separate four-GPU groups under TP4, each served by its own SGLang engine
(Section~\ref{sec:system-implementation}).

\paragraph{Model pairs.}
Task accuracy covers five pairs from three families:
Qwen3 14B$\rightarrow$32B, Qwen3 8B$\rightarrow$14B,
Llama 3.1 8B$\rightarrow$70B, Ministral 3 3B$\rightarrow$14B, and
Ministral 3 8B$\rightarrow$14B.
System efficiency uses Qwen3 14B$\rightarrow$32B. We emphasize
small-to-large handoffs, where avoiding repeated target prefill is most
valuable.
In large-to-small handoffs, four of the five reverse pairs retain
81.46--113.64\% of re-prefill quality on all 11 tasks
(Appendix~\ref{sec:appendix-large-to-small}).
Appendix~\ref{sec:appendix-architecture-extension} adds quality results for
Qwen3.5 4B$\rightarrow$9B and Gemma 3 4B$\rightarrow$12B, and Qwen3.5
serving results.

\paragraph{Baselines.}
\emph{Re-prefill} rebuilds target KV from text history, providing quality
and serving references. Quality and mapper storage compare \method{} with
\closedformmapper, CFM+MLP and
\textsc{MoT} (Section~\ref{sec:background-methods}). Online serving compares \method{}
with re-prefill: \textsc{MoT} is slower than re-prefill at every measured
history length (Section~\ref{sec:motivation-latency}), and
\closedformmapper's serialized handoffs accumulate admission queues
(Appendix~\ref{sec:appendix-efficiency-analysis}). We exclude \textsc{C2C}~\citep{fu2026cache},
which requires target-history computation, and
\textsc{DroidSpeak}~\citep{liu2026droidspeak} and
\textsc{SCD}~\citep{ma2026semantic}, which require matching architectures.
Baseline configurations follow their original papers. CFM and \method{}
use the same 500 FineWeb-Edu sequences for initial fitting, and each
model pair exports one mapper evaluated on all 11 tasks.

\paragraph{Evaluation protocol.}
Thinking mode is disabled throughout. GSM8K retains the reasoning
demonstrations in its eight-shot \texttt{gsm8k\_cot} prompt.
Each handoff splits the source-processed history $H$ from the target query
$Q$, preserving the task-specific scoring or answer-generation continuation.
QuALITY and Qasper test documents and questions are disjoint from those
used to select $k$, and QuALITY additionally excludes all 24 value-map
calibration documents and their associated questions.
Appendix~\ref{sec:appendix-task-protocol} gives the complete prompts,
few-shot counts, metrics and history--query boundaries.

\subsection{System Efficiency}
\label{sec:eval-system}

\paragraph{Agent serving.}
On $\tau$-bench~\citep{yao2024taubench}, Qwen3 14B hands off retail tasks to 32B after
the first successful \texttt{get\_order\_details} or
\texttt{get\_product\_details} call, or after six agent calls.
\method{} succeeds on 43/165 tasks (26.1\%), versus 44 (26.7\%) for
re-prefill.

\begin{figure}[!htbp]
    \centering
    \includegraphics[width=\columnwidth,trim=0 2.25bp 0 1.75bp,clip]{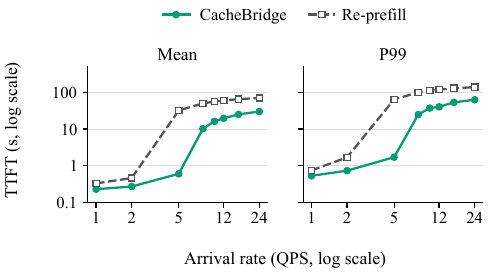}
    \caption{\method{} lowers $\tau$-bench serving TTFT.}
    \label{fig:tau-bench-serving}
\end{figure}

Each rate (1--24 QPS) pools 1,024 requests from two seeds.
Serving TTFT covers waiting, admission, handoff and
target inference through first-token delivery.
Mean TTFT falls by 30--98\% and P99 by 29--97\%
(Figure~\ref{fig:tau-bench-serving}). Appendix~\ref{sec:appendix-efficiency-analysis} details the stage
breakdown and queueing effects.

\begin{figure}[!htbp]
    \centering
    \includegraphics[width=\columnwidth]{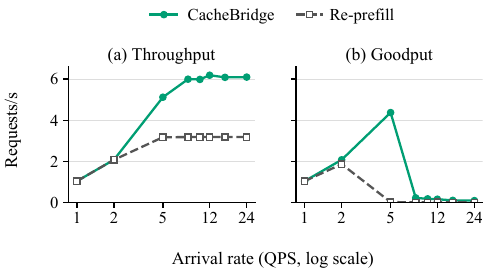}
    \caption{\method{} increases $\tau$-bench throughput and goodput. Goodput requires TTFT $\leq 1$\,s and complete-output latency $\leq 5$\,s.}
    \label{fig:tau-bench-throughput-goodput}
\end{figure}

Throughput divides completions by run duration. Goodput counts requests meeting
TTFT $\leq 1$\,s and complete-output latency $\leq 5$\,s over the same duration.
Throughput reaches 6.18 versus 3.18 requests/s, and goodput at 5 QPS rises
from 0.025 to 4.369 requests/s
(Figure~\ref{fig:tau-bench-throughput-goodput}).
Appendix~\ref{sec:appendix-tau-bench} gives full results, history lengths
and prefill savings.

\paragraph{Long-context and concurrent handoffs.}
Handoff TTFT follows the definition in Section~\ref{sec:motivation-latency},
and at concurrency $C$ requests arrive serially in groups of $C$.
The online serving policy in Section~\ref{sec:system-policy} is disabled
in this experiment.
\begin{figure}[!htb]
    \centering
    \makebox[\columnwidth][c]{%
        \includegraphics[width=\columnwidth,trim=0 3bp 0 3.5bp,clip]{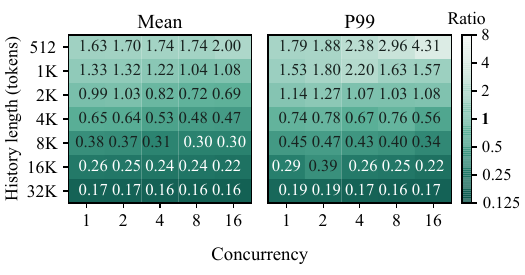}}
    \caption{TTFT ratio of KV transfer to re-prefill with the online serving
    policy in Section~\ref{sec:system-policy} disabled (lower is better).}
    \label{fig:system-ttft}
\end{figure}

At concurrency 1--16 and 4K--32K histories, \method{} reduces mean TTFT
by 35--84\% and P99 by 22--84\% relative to re-prefill,
reaching about 6$\times$ mean speedup at 32K (Figure~\ref{fig:system-ttft}).
Histories of 512--1K favor re-prefill, with a transition at 2K, motivating
length-based routing (Section~\ref{sec:system-policy}).
\paragraph{Distributed handoffs.}
\label{sec:eval-multinode}
We compare \method{} with re-prefill over 512--32K history tokens
and concurrency 1--16 on two servers with TP8 per model
(Section~\ref{sec:system-implementation}). Ratios use the same handoff TTFT
definition as Section~\ref{sec:motivation-latency}.
With source KV prepared before arrival, TTFT is measured on the target
host from handoff arrival to the first host-visible output token, including
all subsequent reconstruction, cross-node communication, installation,
prefix publication, queueing, and query-processing costs.

\begin{figure}[!tbp]
  \centering
  \includegraphics[width=\columnwidth]{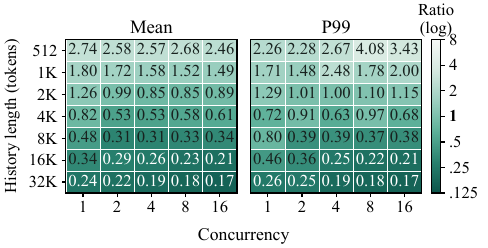}
  \caption{Distributed handoffs improve mean and P99 TTFT at 4K--32K.
  Ratios are \method{}/re-prefill on two servers with TP8 per model
  (lower is better).}
  \label{fig:multinode-tp8-ttft}
\end{figure}

Both mean and P99 TTFT improve at every concurrency for 4K--32K histories
(Figure~\ref{fig:multinode-tp8-ttft}). At 32K and concurrency 16, the ratios
reach 0.167 and 0.169, respectively, about 6$\times$ faster than re-prefill.
Short 512--1K histories favor re-prefill, while 2K lies near the crossover.
Appendix~\ref{sec:appendix-multinode} gives the complete numerical results.

\begin{figure}[H]
    \centering
    \includegraphics[width=\columnwidth]{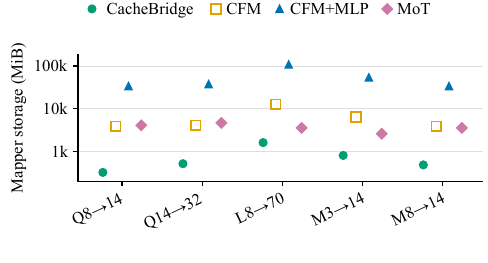}
    \caption{Mapper storage across five model pairs.}
    \label{fig:memory-overhead}
\end{figure}

\paragraph{Memory overhead.}
We report mapper storage separately from temporary handoff memory.
In extra mapper storage (Figure~\ref{fig:memory-overhead}), \method{}
uses 320--1,601\,MiB across five model pairs, about
8--12$\times$ less than CFM and 68--105$\times$ less than CFM+MLP.
Section~\ref{sec:eval-ablation} measures the temporary-memory savings from
direct installation.

\begin{figure*}[!tbp]
    \centering
    \includegraphics[width=\textwidth,trim=0 0.5bp 0 1.5bp,clip]{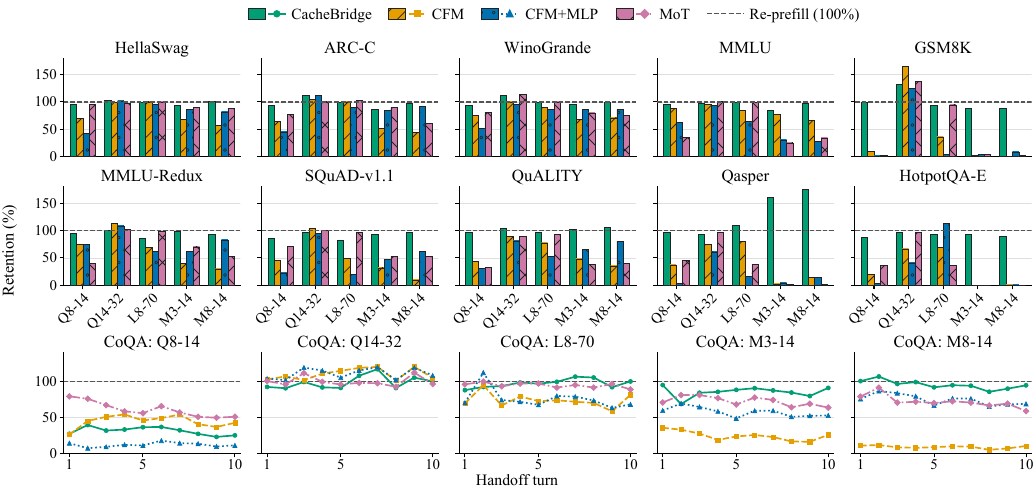}\par
    \caption{\method{} matches or exceeds the strongest transfer baseline in
    38 of 50 single-turn settings.}
    \label{fig:accuracy-results}
\end{figure*}

\subsection{Task Accuracy}
\label{sec:eval-accuracy}

We evaluate chat/reasoning, RAG and multi-turn continuation across five
model pairs. Labels Q, L and M denote Qwen3, Llama 3.1 and Ministral 3,
with sizes in billions. Re-prefill reports 0--100 task scores, and transfer
methods report retention relative to matched re-prefill.
The QuALITY and Qasper test sets (Table~\ref{tab:accuracy-complete-main} in
Appendix~\ref{sec:appendix-complete-accuracy}) are strictly disjoint from
all documents and questions used to select $k$.
QuALITY also excludes the 24 value-map calibration documents and all
associated questions.
Appendix~\ref{sec:appendix-fitting-diagnostics} compares KV fit and
attention-output fidelity.
\method{} achieves the best overall task quality and the most robust
retention among the transfer baselines. Across the 50 single-turn
model-pair/task settings, it matches or exceeds the strongest transfer
baseline in 38 settings, including four ties. Mean retention is 97.94\%
on chat/reasoning and 102.94\% on RAG workloads.
Quality varies across tasks and model pairs.
Appendix~\ref{sec:appendix-qwen-failure} examines shorter continuations and
the quality--storage trade-off of wider reconstruction.

\begin{enumerate}[leftmargin=*,nosep]
    \item \emph{Chat and reasoning.}
    On HellaSwag~\citep{zellers2019hellaswag},
    WinoGrande~\citep{sakaguchi2020winogrande},
    ARC-Challenge~\citep{clark2018arc}, MMLU~\citep{hendrycks2021mmlu},
    MMLU-Redux 2.0~\citep{gema2025mmluredux} and
    GSM8K~\citep{cobbe2021gsm8k}, \method{} matches or exceeds the strongest
    transfer baseline in 22 of 30 settings (Figure~\ref{fig:accuracy-results}).
Retention ranges from 84.09\% to 131.99\%, with 11 settings at or above
re-prefill.

    \item \emph{Context-grounded and RAG tasks.}
    On SQuAD-v1.1~\citep{rajpurkar2016squad}, QuALITY~\citep{pang2022quality},
    Qasper~\citep{dasigi2021qasper} and HotpotQA~\citep{yang2018hotpotqa},
    which cover passage comprehension, long-document multiple choice,
    research-paper QA and multi-hop reasoning, respectively, \method{}
    outperforms the strongest transfer baseline in 16 of 20 settings.
Three pairs each reach or exceed re-prefill on QuALITY and Qasper, and
retention is 83.16--97.36\% on SQuAD and 88.02--96.64\% on HotpotQA-E.

    \item \emph{Multi-turn conversations.}
    On CoQA~\citep{reddy2019coqa}, with one handoff at turns 1--10 scored on
    all later turns, \method{} leads the transfer baselines on three of five
    pairs and retains 85.68--98.87\% of re-prefill quality on four, averaged
    over the handoff positions (Table~\ref{tab:accuracy-complete-coqa-turns}
    in Appendix~\ref{sec:appendix-complete-accuracy}).
    On Qwen3 8B$\rightarrow$14B, an expanded
    mapper raises retention from 31.07\% to 89.11\%
    (Appendix~\ref{sec:appendix-qwen-failure}).

\end{enumerate}

\subsection{Ablation Studies}
\label{sec:eval-ablation}

\paragraph{Compact mapper.}
We substitute CFM's mapper while retaining \method{}'s system
optimizations. Compact mapping lowers mean and P99 TTFT in every
length/concurrency setting, by up to 65\%
(Figure~\ref{fig:cfm-mapper-optimized-path}, with complete ratios in
Table~\ref{tab:cfm-mapper-optimized-path} in
Appendix~\ref{sec:appendix-system-ablation}).

\begin{figure}[!htbp]
  \centering
  \includegraphics[width=\columnwidth]{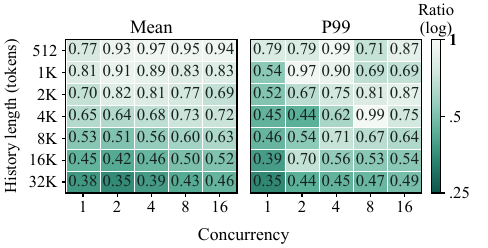}
  \caption{Compact mapping lowers TTFT on the same optimized handoff path.
  Ratios are \method{}/(CFM mapper + \method{} path), lower is better.}
  \label{fig:cfm-mapper-optimized-path}
\end{figure}

\paragraph{Value-map calibration.}
We remove value-map calibration (Section~\ref{sec:fitting-readout}),
retaining the initial key and value maps and head assignment. Across five
model pairs and ten tasks, calibration improves
45 of 50 settings and raises mean retention from 72.26\% to 99.94\%
(Table~\ref{tab:v-calibration-ablation}, with per-task scores in
Table~\ref{tab:accuracy-complete-v-calibration} in
Appendix~\ref{sec:appendix-system-ablation}). The five settings
without improvement are all chat and reasoning tasks.

\begin{table}[H]
    \centering
    \small
    \setlength{\tabcolsep}{3pt}
    \caption{Value-map calibration raises mean retention (\%) for every model pair.}
    \label{tab:v-calibration-ablation}
    \begin{tabular}{@{}lrrrrr@{}}
        \toprule
        Pair & Q8$\to$14 & Q14$\to$32 & L8$\to$70 & M3$\to$14 & M8$\to$14 \\
        \midrule
        w/o V cal. & 55.89 & 93.89 & 37.31 & 83.29 & 90.94 \\
        Full & 94.09 & 104.89 & 96.23 & 99.61 & 104.90 \\
        \bottomrule
    \end{tabular}
\end{table}

\paragraph{Execution components.}
Each ablation disables one component on Qwen3 14B$\rightarrow$32B,
fixing source-side mapping computation placement, mapper, TP, block size,
workload and memory cap. Direct installation and pipeline overlap test
Section~\ref{sec:system-concurrency}; path scheduling, request concurrency
and dynamic buffer allocation test Section~\ref{sec:system-policy}.
Sections~\ref{sec:motivation-support} and~\ref{sec:eval-system} quantify
what source-side reconstruction avoids.
Appendix~\ref{sec:appendix-system-ablation} gives detailed results.

\noindent\emph{Direct installation.}
We compare direct installation with staged copy, using 512 requests per
history length at concurrency 1. At 32K, direct installation reduces peak added target memory
by 99.6\% and median TTFT by 1.0\%
(Table~\ref{tab:direct-installation}).

\begin{table}[H]
\centering
\small
\setlength{\tabcolsep}{3pt}
\caption{Direct installation lowers TTFT and added target memory.}
\label{tab:direct-installation}
\begin{tabular}{@{}lrrrr@{}}
\toprule
 & \multicolumn{2}{c}{Median TTFT (ms)} & \multicolumn{2}{c}{Memory (MiB)} \\
\cmidrule(lr){2-3}\cmidrule(l){4-5}
History & Direct & Copy & Direct & Copy \\
\midrule
2K & 87.36 & 88.74 & 5.56 & 128.00 \\
8K & 126.88 & 128.83 & 7.83 & 512.00 \\
32K & 293.39 & 296.35 & 8.04 & 2,048.00 \\
\bottomrule
\end{tabular}
\end{table}

\noindent\emph{Pipeline overlap.}
We serialize reconstruction, transfer and installation per request,
retaining direct installation.
\begin{samepage}
\mbox{Overlap reduces} mean TTFT by 12--35\%
across all configurations and P99 TTFT by 5--27\% at 32K
(Figure~\ref{fig:pipeline-overlap}).

\nopagebreak[4]
\begin{figure}[H]
    \centering
    \includegraphics[width=\columnwidth,trim=0 1.5bp 0 0.75bp,clip]{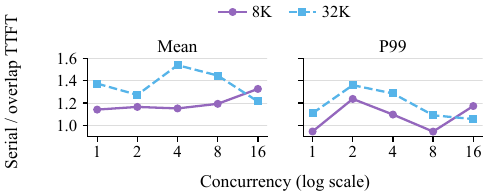}
    \caption{Overlap reduces mean TTFT.}
    \label{fig:pipeline-overlap}
\end{figure}

\end{samepage}

\begin{samepage}
\noindent\emph{Path scheduling.}
We force transfer for 512 requests equally mixing 512-token, 1K, 2K, 4K
and 8K histories, retaining memory admission and re-prefill fallback.
Path scheduling reduces mean TTFT by 20--63\% and P99 TTFT by 23--71\%
(Figure~\ref{fig:path-scheduling}).

\nopagebreak[4]
\begin{figure}[H]
    \centering
    \includegraphics[width=\columnwidth,trim=0 1.5bp 0 1.5bp,clip]{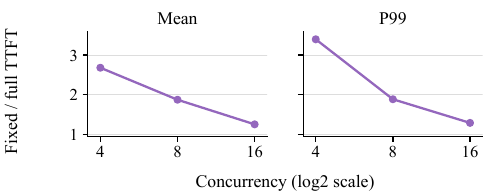}
    \caption{Path scheduling lowers TTFT.}
    \label{fig:path-scheduling}
\end{figure}

\end{samepage}

\begin{figure}[!htbp]
    \centering
    \includegraphics[width=\columnwidth,trim=0 1.5bp 0 0.75bp,clip]{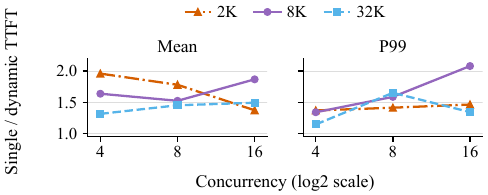}
    \caption{Request concurrency lowers TTFT.}
    \label{fig:request-concurrency}
\end{figure}

\noindent\emph{Request concurrency.}
We limit admitted transfer pipelines to one under the same offered load.
Request concurrency reduces mean TTFT by 24--49\%
and P99 TTFT by 13--52\% across all configurations
(Figure~\ref{fig:request-concurrency}).

\noindent\emph{Dynamic buffer allocation.}
On the mixed-history workload of the path-scheduling ablation, we fix the
pool at two buffer sets (channels) under the same memory cap.
Dynamic allocation reduces mean TTFT by 21\%, P99 TTFT by 14\% and
average pool residency by 22\% (Table~\ref{tab:dynamic-buffer-allocation}).

\begin{table}[!t]
\centering
\small
\setlength{\tabcolsep}{3pt}
\caption{Dynamic allocation lowers TTFT and average pool residency.}
\label{tab:dynamic-buffer-allocation}
\begin{tabular}{@{}lrrr@{}}
\toprule
 & \multicolumn{2}{c}{TTFT (ms)} & Avg.\ pool \\
\cmidrule(lr){2-3}
Pool & Mean & P99 & (MiB/rank) \\
\midrule
Dynamic & 729.52 & 1,503.52 & 526.38 \\
Fixed (2 channels) & 923.42 & 1,739.26 & 672.00 \\
\bottomrule
\end{tabular}
\end{table}

\section{Related Work}
\label{sec:related}

To our knowledge, \method{} is the first serving system to accelerate
heterogeneous multi-model workloads through cross-model KV transfer.

\paragraph{Cross-model KV transfer.}
\closedformmapper fits affine maps~\citep{heo2026crossmodel}, and latent
alignment enables cache communication~\citep{dery2026latentalign}.
DroidSpeak and Activated LoRA reuse caches among base-model
variants~\citep{liu2026droidspeak,li2025alora}. A context-reuse layer studies
handoffs across scales, architectures, tokenizers and families, focusing
on quality and prefill cost~\citep{li2026universal}.

\paragraph{LLM serving.}
Serving systems optimize cache
management~\citep{kwon2023pagedattention,zheng2023sglang,yao2025cacheblend,liu2024cachegen,qin2024mooncake,qu2026haserve}
and separate prefill from decode~\citep{zhong2024distserve}. Multi-model
systems multiplex or load
models~\citep{li2023alpaserve,duan2024muxserve,fu2024serverlessllm} and batch
LoRA adapters~\citep{sheng2024slora,chen2024punica,wu2024dlora}, while
reusing KV within each model.

\section{Limitations}
\label{sec:limitations}

\paragraph{Model-pair sensitivity.}
Transfer quality varies across source--target model pairs, even within the
same LLM family. This sensitivity is particularly pronounced in multi-turn
continuation. Appendix~\ref{sec:appendix-qwen-failure} recovers quality on
Qwen3 8B$\rightarrow$14B with an expanded mapper at greater storage cost.

\paragraph{Within-family transfer.}
\method{} supports KV transfer between models from the same LLM family
that share a tokenizer (Section~\ref{sec:fitting-interface}). Extending
transfer across model families remains future work.

\section{Conclusion and Outlook}
\label{sec:conclusion}

We presented \method{}, the first serving system to accelerate
heterogeneous multi-model workloads through cross-model KV transfer. It
co-designs a compact, attention-calibrated mapper with the handoff path to
preserve task quality, reduce latency and memory overhead, and improve
throughput.

Looking ahead, multi-model LLM serving with cross-model KV transfer
introduces a workload in which model selection also determines how
conversation state can be reused. This opens opportunities to jointly
optimize model routing, cache retention and handoff scheduling across
turns, balancing quality, latency and memory over complete multi-model
workflows.

\label{main:body-end}

\clearpage
\bibliographystyle{configuration/plain-linked}
\bibliography{resources/reference}

\clearpage
\appendix
\setlength{\multicolsep}{2pt}
\raggedbottom
\setcounter{topnumber}{4}
\setcounter{bottomnumber}{4}
\setcounter{totalnumber}{6}
\setcounter{dbltopnumber}{3}
\renewcommand{\topfraction}{0.95}
\renewcommand{\bottomfraction}{0.9}
\renewcommand{\textfraction}{0.05}
\renewcommand{\floatpagefraction}{0.85}
\renewcommand{\dbltopfraction}{0.95}
\renewcommand{\dblfloatpagefraction}{0.85}
\makeatletter
\setlength{\@fptop}{0pt}
\setlength{\@dblfptop}{0pt}
\setlength{\@fpsep}{12pt plus 2pt minus 2pt}
\setlength{\@dblfpsep}{12pt plus 2pt minus 2pt}
\makeatother
\section{Evaluation and Implementation Details}
  \label{sec:appendix-task-protocol}
  \paragraph{Evaluation coverage and decoding.}
  For QuALITY, the evaluation set excludes the 24 calibration documents
  and all associated questions. Other tasks use their complete evaluation sets.
  Table~\ref{tab:task-evaluation-protocol} specifies each task's prompt,
  few-shot count and primary metric. Retention divides each transfer score
  by the matched re-prefill score, so values above 100\% mean that transfer
  scores higher; Table~\ref{tab:accuracy-complete-main} gives the raw
  re-prefill scores. Thinking mode is disabled in all evaluations. GSM8K retains the reasoning demonstrations in its
  eight-shot \texttt{gsm8k\_cot} prompt. For CoQA, every handoff position
  is evaluated on the complete evaluation set.

  \paragraph{Baseline configuration.}
  Baseline configurations follow their respective original papers,
  including the source-layer selection of CFM and CFM+MLP and the training
  setups of CFM+MLP and \textsc{MoT}.
  CFM and \method{} use the same initial fitting data, namely the
  500 FineWeb-Edu sequences described in
  Appendix~\ref{sec:appendix-calibration}.
  Each model pair exports one mapper, which is evaluated on all 11 tasks.

\begin{table*}[!b]
\centering
\begingroup
\fontsize{8.5}{10.2}\selectfont
\setlength{\tabcolsep}{4pt}
\renewcommand{\arraystretch}{1.0}
\caption{Task prompts, few-shot counts and primary scoring metrics. SQuAD and CoQA also report EM as an auxiliary metric. Thinking mode is disabled throughout.}
\label{tab:task-evaluation-protocol}
\begin{tabular}{@{}p{0.14\textwidth}p{0.43\textwidth}cp{0.32\textwidth}@{}}
\toprule
Task & Prompt setting & Shots & Primary metric \\
\midrule
HellaSwag & Original lm-eval completion template, comparing candidate continuations & 0 & Length-normalized option accuracy \\
ARC-C & Original lm-eval question--answer template, comparing candidate answer continuations & 0 & Length-normalized option accuracy \\
WinoGrande & Original lm-eval cloze-completion template & 0 & Option accuracy \\
MMLU & lm-eval subject-specific multiple-choice template, comparing option continuations & 5 & Option accuracy \\
GSM8K & lm-eval \texttt{gsm8k\_cot}, with a newline between \texttt{Q:}\ldots\ and \texttt{A:}, and reasoning demonstrations & 8 & EM after numeric-answer extraction \\
MMLU-Redux 2.0 & Question and A--D options, requesting only the answer letter & 0 & Accuracy after answer-letter parsing \\
SQuAD-v1.1 & \texttt{Context:}\ldots\ followed by an instruction to return a short answer using only the context & 0 & Token F1 \\
QuALITY & Article, question and A--D options, requesting only the best answer letter & 0 & Option accuracy from A--D single-token logits \\
Qasper & Full paper, requesting a short answer and supporting Yes/No and Unanswerable & 0 & Official Answer F1 \\
HotpotQA-E & Given passages, requesting only the answer & 0 & English QA token F1, maximized over references \\
CoQA & Passage, previous question--answer history and current question, requesting a short answer & 0 & CoQA token F1 \\
\bottomrule
\end{tabular}
\endgroup
\end{table*}

\paragraph{History--query boundary.}
  Table~\ref{tab:task-handoff-splits} specifies the source history $H$
  and the target query $Q$ for the single-turn tasks.
  Each split fixes the source-processed prefix while preserving the
  target's task-specific scoring or answer-generation continuation.
  For CoQA, the prompt contains the passage, previous question--answer
  history and current question, with the turn-based handoff and scoring
  scope specified in Section~\ref{sec:eval-accuracy}.

\begin{table*}[!tbp]
\centering
\begingroup
\fontsize{8.5}{10.2}\selectfont
\setlength{\tabcolsep}{4pt}
\renewcommand{\arraystretch}{1.0}
\caption{Source-history and target-query splits for the single-turn tasks. Teacher forcing scores the candidate continuation after the target processes the final context token.}
\label{tab:task-handoff-splits}
\begin{tabular}{@{}>{\raggedright\arraybackslash}p{0.19\textwidth}>{\raggedright\arraybackslash}p{0.385\textwidth}>{\raggedright\arraybackslash}p{0.385\textwidth}@{}}
\toprule
Task & Source history $H$ & Target query $Q$ \\
\midrule
HellaSwag, ARC-C, WinoGrande, MMLU & Complete context for each candidate except its last token, including the current question and any few-shot demonstrations & Last context token, followed by the teacher-forced candidate continuation \\
GSM8K & Complete eight-shot prompt except its last token, including the current question & Last prompt token, followed by answer generation \\
MMLU-Redux 2.0 & Task description, current question and all options & Instruction to output only A/B/C/D, followed by \texttt{Answer:} \\
SQuAD-v1.1 & \texttt{Context:}, a newline, the passage and two trailing newlines & Short-answer instruction, current question and \texttt{Answer:} \\
QuALITY & Short-answer instruction and complete article & Current question, four options, \texttt{Answer:} and template suffix \\
Qasper & Short-answer instruction, paper title, abstract and all section bodies & Current question, \texttt{Answer:} and template suffix \\
HotpotQA-E & Passages and associated instructions, ending before the current \texttt{Question:} & Current question, \texttt{Answer:} and template suffix \\
\bottomrule
\end{tabular}
\endgroup
\end{table*}

\paragraph{Implementation details.}
\label{sec:appendix-implementation}
Corresponding source and target ranks exchange CUDA IPC memory handles and
ready/ACK messages over Unix-domain sockets. Each target rank copies
reconstructed blocks from imported source slots into its unpublished KV
pages and acknowledges each copy to release the slot. The Python
coordinator communicates with the engines over multiprocessing pipes.
Engines relay control operations to scheduler ranks through SGLang's
ZeroMQ collective-RPC interface. Bounded request queues apply backpressure,
with a separate control queue for acknowledgments.

\section{Calibration Configuration}
\label{sec:appendix-calibration}

\paragraph{Initial K/V fitting.}
The compact mapper starts from an affine ridge fit, following
\closedformmapper~\citep{heo2026crossmodel}.
It uses 500 FineWeb-Edu~\citep{penedo2024fineweb} sequences of 1,024 tokens
each, with a token-sampling stride of four.
This yields 128,000 token positions for fitting.
Both frozen models process the same sequences, and source and target KV
vectors are paired by sequence and token position. Key regressions use
content coordinates, with source and target RoPE rotations removed.
For each source layer $j$, target layer $\ell$ and matched KV head $h$, we
fit centered single-layer linear regressions separately for K and V. Their
coefficients of determination give the source-layer score
\begin{equation}
  s_{j,\ell}=\frac{1}{2H_s}\sum_{h=1}^{H_s}
    \left(R^2_{K,j,\ell,h}+R^2_{V,j,\ell,h}\right),
  \label{eq:fitting-layer-score}
\end{equation}
and $\mathcal S_\ell$ contains the top-$k$ source layers by this score.
For each target layer, KV head and component, let $X_c,Y_c$ be centered
source features and re-prefill targets, and $\bar X,\bar Y$ their
pre-centering means. Initial affine ridge gives
\begin{equation}
  \begin{split}
    W^0&=(X_c^{\mathsf T}X_c+\lambda I)^{-1}X_c^{\mathsf T}Y_c,\\
    b^0&=\bar Y-\bar XW^0,
  \end{split}
  \label{eq:fitting-native}
\end{equation}
where $I$ is the identity and $\lambda>0$ regularizes the fit.
Each solve is independent across target layers, KV heads and components.
The initial affine ridge solver consumes sufficient statistics. We gather
the selected source-head features in bounded panels and accumulate the
centered statistics for head-batched solves.
For all five model pairs, we set the initial ridge coefficient to
$\lambda=0.01$, matching the setting in
\closedformmapper~\citep{heo2026crossmodel}.
Table~\ref{tab:lambda-sensitivity-hellaswag} examines sensitivity to this
coefficient on HellaSwag for Qwen3 14B$\rightarrow$32B.
Across $\lambda=0$--$100$, \method{} retains 96.61--104.00\% of re-prefill
task quality, while CFM ranges from 66.32\% to 98.75\%, showing that
\method{} is less sensitive to the initial ridge coefficient.

\begin{table}[!tbp]
    \centering
    \small
    \setlength{\tabcolsep}{7pt}
    \caption{Sensitivity to the initial ridge coefficient $\lambda$ on
    HellaSwag for Qwen3 14B$\rightarrow$32B. Both methods report task-quality
    retention relative to re-prefill. The adopted $\lambda=0.01$ is bold.}
    \label{tab:lambda-sensitivity-hellaswag}
    \begin{tabular}{@{}rrr@{}}
        \toprule
        $\lambda$ & CFM (\%) & \method{} (\%) \\
        \midrule
        0 & 97.83 & 97.71 \\
        0.0001 & 97.86 & 97.09 \\
        \textbf{0.01} & \textbf{98.75} & \textbf{104.00} \\
        0.1 & 96.49 & 97.09 \\
        1 & 78.57 & 96.61 \\
        100 & 66.32 & 97.01 \\
        \bottomrule
    \end{tabular}
\end{table}

\begin{figure}[!htbp]
    \centering
    \includegraphics[width=\columnwidth]{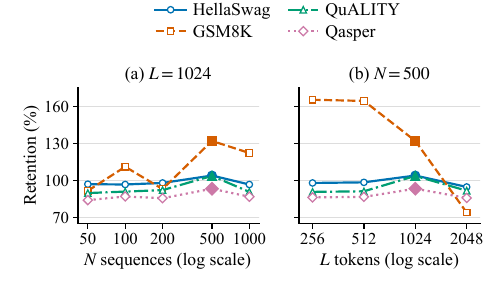}
    \caption{GSM8K is more sensitive to fitting-data size than the other three tasks. Qwen3 14B$\rightarrow$32B retention relative to re-prefill, varying sequence count $N$ or token length $L$. Filled markers show the shared default $(N,L)=(500,1024)$.}
    \label{fig:fitting-data-sensitivity}
\end{figure}

\paragraph{Fitting-data sensitivity.}
We vary the number $N$ of initial-fitting sequences and their token length
$L$ on Qwen3 14B$\rightarrow$32B (Figure~\ref{fig:fitting-data-sensitivity}).
The $N$ sweep fixes $L=1024$, and the $L$ sweep fixes $N=500$.
The four evaluation datasets cover commonsense completion (HellaSwag),
multi-step arithmetic reasoning (GSM8K), long-document comprehension
(QuALITY) and information-seeking QA over research papers (Qasper), spanning
reasoning and context-grounded tasks.
The default $(N,L)=(500,1024)$ appears in both sweeps.
It yields the highest measured retention on all four tasks in the $N$
sweep and on HellaSwag, QuALITY and Qasper in the $L$ sweep.

\begin{table}[!tbp]
    \centering
    \small
    \setlength{\tabcolsep}{7pt}
    \caption{Number of selected source layers $k$ for initial KV fitting.}
    \label{tab:calibration-support}
    \begin{tabular}{@{}lr@{}}
        \toprule
        Model pair & $k$ \\
        \midrule
        Qwen3 8B$\rightarrow$14B & 8 \\
        Qwen3 14B$\rightarrow$32B & 8 \\
        Llama 3.1 8B$\rightarrow$70B & 20 \\
        Ministral 3 3B$\rightarrow$14B & 20 \\
        Ministral 3 8B$\rightarrow$14B & 12 \\
        \bottomrule
    \end{tabular}
\end{table}

\paragraph{Layer budget and online trade-off.}
We fix one $k$ per model pair across tasks
(Table~\ref{tab:calibration-support}).
We choose $k$ on independent QuALITY and Qasper validation data, balancing
quality against mapper storage and reconstruction work. These validation
documents and all associated questions are strictly disjoint from the
final test sets in Table~\ref{tab:accuracy-complete-main}.
Table~\ref{tab:k-sensitivity} reports sensitivity on the test sets after
this choice is fixed, with the validation-selected $k$ in bold.
Both Qwen3 pairs attain their highest measured retention on these
tasks at $k=8$, and the Llama pair peaks at $k=20$.
The Ministral results vary by task and are non-monotonic as $k$ grows;
their adopted $k$ stays within three percentage points of the best
measured retention on both tasks.

\begin{table*}[!tbp]
\centering
\begingroup
\fontsize{8.5}{10.2}\selectfont
\setlength{\tabcolsep}{4pt}
\renewcommand{\arraystretch}{1.0}
\caption{Task-quality retention (\%) versus selected source layers $k$. Bold entries use the adopted $k$.}
\label{tab:k-sensitivity}
\begin{tabular*}{\textwidth}{@{\extracolsep{\fill}}lrrrrrrrr@{}}
\toprule
Task & $k=1$ & $k=2$ & $k=4$ & $k=8$ & $k=12$ & $k=16$ & $k=20$ & $k=24$ \\
\midrule
\multicolumn{9}{@{}l}{\textbf{Qwen3 14B$\rightarrow$32B}} \\*
QuALITY & 88.42 & 89.13 & 89.83 & \textbf{103.70} & 92.43 & 91.02 & 91.96 & 91.49 \\
Qasper & 84.08 & 86.52 & 86.71 & \textbf{93.34} & 88.50 & 86.85 & 89.20 & 86.86 \\
\midrule
\multicolumn{9}{@{}l}{\textbf{Qwen3 8B$\rightarrow$14B}} \\*
QuALITY & 54.61 & 61.35 & 71.07 & \textbf{97.09} & 84.29 & 85.04 & 88.28 & 87.28 \\
Qasper & 42.77 & 44.92 & 52.55 & \textbf{96.78} & 62.70 & 67.63 & 67.79 & 71.48 \\
\midrule
\multicolumn{9}{@{}l}{\textbf{Llama 3.1 8B$\rightarrow$70B}} \\*
QuALITY & 44.25 & 52.08 & 56.48 & 56.97 & 59.66 & 64.30 & \textbf{96.49} & 69.68 \\
Qasper & 30.92 & 36.62 & 38.68 & 38.88 & 41.75 & 69.17 & \textbf{109.75} & 8.30 \\
\midrule
\multicolumn{9}{@{}l}{\textbf{Ministral 3 3B$\rightarrow$14B}} \\*
QuALITY & 89.53 & 91.57 & 96.80 & 100.29 & 100.58 & 102.33 & \textbf{102.03} & 102.62 \\
Qasper & 108.86 & 115.14 & 124.68 & 141.48 & 155.97 & 157.45 & \textbf{160.93} & 163.65 \\
\midrule
\multicolumn{9}{@{}l}{\textbf{Ministral 3 8B$\rightarrow$14B}} \\*
QuALITY & 101.74 & 102.62 & 104.36 & 105.23 & \textbf{106.98} & 105.81 & 104.98 & 104.56 \\
Qasper & 153.09 & 166.21 & 175.30 & 176.35 & \textbf{175.18} & 176.46 & 175.02 & 171.65 \\
\bottomrule
\end{tabular*}
\endgroup
\end{table*}

For a fixed model pair, each map has $kd_s$ input features, so increasing
$k$ increases its non-bias coefficient storage and affine
multiply--add count linearly with $k$.
It also expands the source features packed for reconstruction.
These costs consume source-side compute and memory during serving and
compete with concurrent handoffs.
Under source-side reconstruction, each block still sends only target KV,
whose dimensions are fixed by the target model and block size.
Thus, larger $k$ raises local reconstruction cost while the transmitted
target-KV payload and output-buffer dimensions remain unchanged
(Section~\ref{sec:system-placement}).

Table~\ref{tab:k-latency} measures this online trade-off on Qwen3
14B$\rightarrow$32B at concurrency 1.
At 8K history, increasing $k$ from 8 to 24 raises handoff TTFT from
158.9 to 205.4\,ms (29.3\%) without improving retention on either
task in Table~\ref{tab:k-sensitivity}.
The adopted $k=8$ combines the highest measured retention on both tasks
with 1.60$\times$ and 2.51$\times$ speedups over re-prefill at 4K and
8K histories, respectively.
Re-prefill remains faster at 512--1K across all tested $k$, while
2K lies near the crossover, supporting length-based path selection.

\begin{table}[!tbp]
\centering
\begingroup
\fontsize{8.5}{10.2}\selectfont
\setlength{\tabcolsep}{3.5pt}
\renewcommand{\arraystretch}{1.08}
\caption{Handoff TTFT (ms) versus $k$ and history length on Qwen3 14B$\rightarrow$32B at concurrency 1. The adopted $k=8$ is bold.}
\label{tab:k-latency}
\begin{tabular}{@{}lrrrrr@{}}
\toprule
$k$ / path & 512 & 1K & 2K & 4K & 8K \\
\midrule
Re-prefill & 77.7 & 88.2 & 122.9 & 205.2 & 399.3 \\
\midrule
1 & 108.8 & 109.4 & 113.2 & 117.3 & 146.2 \\
2 & 106.8 & 123.3 & 122.4 & 118.6 & 138.7 \\
4 & 112.1 & 109.0 & 110.8 & 125.5 & 144.5 \\
\textbf{8} & \textbf{112.6} & \textbf{117.7} & \textbf{118.9} & \textbf{128.1} & \textbf{158.9} \\
12 & 115.7 & 119.0 & 120.6 & 134.2 & 164.5 \\
16 & 115.1 & 127.6 & 121.8 & 147.1 & 168.2 \\
20 & 122.7 & 115.6 & 126.4 & 150.2 & 193.6 \\
24 & 125.0 & 128.2 & 124.0 & 157.4 & 205.4 \\
\bottomrule
\end{tabular}
\endgroup
\end{table}

\paragraph{Value-map calibration with attention readouts.}
The second stage uses 24 documents from QuALITY with four queries per document, giving
96 complete query sequences for the layer-ordered value-map correction in
Section~\ref{sec:fitting-readout}. For each document--query pair $(H,q)$,
the source processes $H$, the initial mapper reconstructs target history
KV, and the target processes the appended query $q$. History KV from
re-prefill provides the reference at the current query state. These
documents and all associated questions are excluded from QuALITY evaluation.

\paragraph{Readout weighting.}
For a fixed target layer and KV head $h$, let $\mathcal G(h)$ contain its
$g$ query heads in grouped-query attention (GQA). Query $q_{di}$ is the
complete $i$th query for document $d$ and has $m_{di}$ tokens. The matrices
$D_{q_{di}}$ and $R_{q_{di}}$ stack all token rows from these $g$ query
heads. Let $a_{diut}$ and $r_{diut}$ denote their rows for query head $u$
and token $t$. Setting $w_{q_{di}}=w_{di}=1/(4g m_{di})$, the objective in
Equation~\ref{eq:fitting-readout} expands to
\begin{equation}
  \begin{aligned}
    \min_{\Delta\Theta}\quad&
      \sum_{d=1}^{24}\sum_{i=1}^{4}w_{di}
      \sum_{u\in\mathcal G(h)}\sum_{t=1}^{m_{di}}
      \|a_{diut}\Delta\Theta-r_{diut}\|_2^2\\
      &\quad+\lambda_{\mathrm{ro}}\|\Delta\Theta\|_F^2\\
    \text{subject to}\quad&
      \operatorname{tr}(\Delta\Theta^{\mathsf T}M\Delta\Theta)
      \leq 0.01E_0.
  \end{aligned}
  \label{eq:fitting-readout-expanded}
\end{equation}
Within each query, we average over tokens and its GQA heads and sum
squared errors across output coordinates. Each document contributes the
mean loss of its four queries. Summing these document losses gives equal
weight to the 24 documents.

\paragraph{Sufficient statistics and ridge.}
We accumulate
\begin{equation}
  \begin{aligned}
    G&=\sum_{d,i,u,t}w_{di}\,a_{diut}^{\mathsf T}a_{diut},\\
    C&=\sum_{d,i,u,t}w_{di}\,a_{diut}^{\mathsf T}r_{diut}.
  \end{aligned}
  \label{eq:fitting-readout-statistics}
\end{equation}
The index ranges follow Equation~\ref{eq:fitting-readout-expanded}, so
both statistics retain the sum over documents. We use isotropic ridge
with
\begin{equation}
  \lambda_{\mathrm{ro}}
    =\max\!\left(10^{-8},\,0.01\,\frac{\operatorname{tr}(G)}{p}\right),
  \label{eq:fitting-readout-ridge}
\end{equation}
where $p$ is the readout-feature width, including the intercept.

\paragraph{Energy budget and constrained solve.}
The update space comprises the existing value-map weights and bias for each
target KV head. We measure the source-feature second moment and the
initial mapper's value residual energy relative to re-prefill on the same
4,096 sampled initialization states. Their token averages define the
metric $M$ and reference energy $E_0$ for each layer and KV head.
The source-feature second moment includes the intercept and $10^{-4}$
diagonal stabilization. The energy budget is $0.01E_0$, and the
history-mass-pooled intercept shares both the ridge penalty and the
energy constraint.

The constrained solution has the form
\begin{equation}
  \Delta\Theta(\mu)
    =(G+\lambda_{\mathrm{ro}}I+\mu M)^{-1}C,
  \label{eq:fitting-readout-solution}
\end{equation}
where $\mu\geq0$ enforces the energy budget. We use $\mu=0$ when the
unconstrained solution satisfies the budget. Otherwise, a scalar search
selects $\mu$ at the energy boundary. Each head is solved in FP64.
Folding retains the original coefficient dtype without additional
clipping or projection.

\section{Model Architecture Extension}
\label{sec:appendix-architecture-extension}

\subsection{From Full Attention to Mixed Layer Types}

Recent models reduce long-context memory and computation by using local
attention or combining attention with recurrent layers.
A Transformer with full attention stores a key and value for each history
token at every layer.
Gemma 3 instead interleaves five local-attention layers with one
global-attention layer. Its local layers attend to a 1,024-token window,
while its global layers retain access to the full
history~\citep{gemmateam2025gemma3}.
Qwen3.5 4B and 9B combine three Gated DeltaNet layers with one gated
attention layer in each repeated block~\citep{qwen2026qwen35models}.
The attention layers store token-level KV, whereas Gated DeltaNet also
requires a recurrent memory matrix and a short convolution buffer.
Both evaluated families use dense feed-forward networks. The distinction
relevant to transfer is the type of sequence layer and the state it
retains.

We use Gemma 3 4B$\rightarrow$12B and Qwen3.5 4B$\rightarrow$9B to examine
how the fitting workflow in Section~\ref{sec:fitting} accommodates these
changes.

\subsection{Reusing the Offline Fitting Workflow}

\method{} collects source and target states from the same unlabeled
calibration text, selects source layers and heads that match the
target's structure, and fits their coefficients with both models frozen.
For these extensions, ridge regression fits the state maps, and a
rank-constrained spectral decomposition retains a limited number of
directions for the residual correction. These are closed-form fitting
steps and require no gradient training.
A new model pair needs newly calibrated coefficients, while the data
collection and fitting procedure remain the same.

The fitting targets follow the state required for continuation.
For an attention layer, the targets are its history keys and values.
A recurrent layer adds its memory at the handoff position.
A convolution layer also needs the recent inputs retained in its buffer.
Table~\ref{tab:architecture-fitting} summarizes the resulting changes.
It also outlines a further extension to RecurrentGemma, which combines
local attention with a Real-Gated Linear Recurrent Unit
(RG-LRU)~\citep{botev2024recurrentgemma}.

\begin{table*}[!tbp]
    \centering
    \small
    \setlength{\tabcolsep}{4pt}
    \caption{Continuation states and fitting adjustments across architectures.}
    \label{tab:architecture-fitting}
    \begin{tabular}{@{}p{0.19\textwidth}p{0.32\textwidth}p{0.45\textwidth}@{}}
        \toprule
        Architecture & State required for continuation & Fitting adjustment \\
        \midrule
        Gemma 3 & KV for local and global attention, with the target's
        KV-head grouping and window sizes & Select compatible layer types,
        assign source heads to target heads, and fit KV in the target's
        normalization and RoPE conventions. \\
        \addlinespace
        Qwen3.5 & Attention KV, Gated DeltaNet memory matrices, and
        convolution buffers & Fit attention maps by regression. Transform
        each memory matrix along its key and value dimensions, and
        reconstruct convolution history separately. \\
        \addlinespace
        RecurrentGemma\newline (further extension) & Local-attention KV,
        RG-LRU recurrent vectors, and convolution buffers & Fit a linear
        readout for the recurrent vector, with separate convolution-buffer
        and handoff-position handling. \\
        \bottomrule
    \end{tabular}
\end{table*}

\paragraph{Attention maps.}
The extension retains a head-local base map and adds a rank-limited
residual fitted from source K/V features.
The base provides a compact correspondence between heads. The residual
supplies information that this correspondence misses.
The final V fit holds the selected source-V subspace fixed and solves
jointly for its coefficients and a contribution from source K.
Because the residual can use other source heads, the complete map is a
structured mapping with a head-local base.
Its additional source dependencies are recorded with the exported mapper.

Calibration follows the target's computation.
Normalization determines the coordinates being fitted. Local-attention
masks determine which history positions are visible. Gates and the output
readout determine which reconstruction errors affect the layer output.
These operations guide which directions the limited residual capacity
retains. For Gemma 3, we also select source layers and heads within compatible
local or global layers and restore the target's RoPE and cache-window
layout.

\paragraph{Qwen3.5 memory matrices.}
For each recurrent head, calibration fits one linear transformation for
key features and another for value features. Call these transformations
$A$ and $B$.
Writing a source memory matrix $U$ with its key dimension along rows and
value dimension along columns, the corresponding target estimate is
\begin{equation}
    \widehat U = A^{\mathsf T} U B.
    \label{eq:architecture-memory-map}
\end{equation}
Thus the same fitted transformations act on the two dimensions of the
memory. When several compatible source layers are selected, we average
their transformed memory estimates.
This reconstructs a target state approximately; subsequent updates
still follow the target's own gates and recurrent dynamics.
The convolution buffer is reconstructed separately.
Attention-readout calibration applies to the attention layers; recurrent
memory requires its own state targets and continuation checks.

\subsection{Reconstructing State at the Handoff}

Fitting and target computation have separate roles at a handoff.
The continuation procedure consists of three steps:
\begin{enumerate}[leftmargin=*,nosep]
    \item Map the source history state to target KV and, where required,
    recurrent memory and convolution history.
    \item Arrange the mapped state in the target's cache layout and
    perform the required first-layer reconstruction.
    \item Run the target on the configured suffix of history tokens
    before processing the new query.
\end{enumerate}
For a suffix of $r$ tokens, the source state is taken immediately before
that suffix, so replay advances the reconstructed state to the handoff
position. This target computation is accounted for separately from
the fitted map. For Gemma 3 4B$\rightarrow$12B, the head maps retain their
closed-form coefficients, and the target adds first-layer reconstruction
and replays the last $r=\min(64,\lfloor H/4\rfloor)$ tokens of an
$H$-token history.
Online cost includes state mapping, installation, first-layer
reconstruction, suffix replay, and query processing through the first
target token.

\subsection{System Efficiency}
\label{sec:appendix-qwen35-efficiency}

\begin{figure}[!tbp]
    \centering
    \includegraphics[width=\columnwidth]{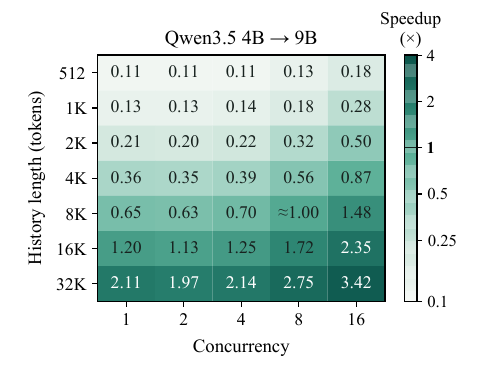}
    \caption{The architecture-adapted Qwen3.5 path retains TTFT gains at
    long histories. Cells show re-prefill mean TTFT divided by
    \method{} mean TTFT, so values above 1 favor \method{}.
    The color scale is logarithmic, with darker green indicating larger
    speedups. The 8K, $C=8$ point is approximately 1.00.}
    \label{fig:qwen35-system-speedup}
\end{figure}

Figure~\ref{fig:qwen35-system-speedup} evaluates Qwen3.5
4B$\rightarrow$9B across 512--32K history tokens and concurrency 1--16.
Replacing full attention with a mixture of attention and recurrent layers
changes both the target's re-prefill work and the state reconstructed at
handoff. The adapted path handles attention KV, recurrent memory and
convolution buffers, together with the continuation work described above.
These changes affect the history length at which transfer becomes
beneficial and the size of the serving gain.

Compared with the full-attention Qwen3 pair in
Section~\ref{sec:eval-system}, Qwen3.5 has a later crossover and smaller
long-history speedups. Re-prefill is faster at 512--4K for all tested
concurrency levels. At 8K, \method{} reaches approximate parity at
$C=8$ and a 1.48$\times$ speedup at $C=16$.
At 16K and 32K, transfer benefits every tested concurrency level,
with speedups of 1.13--2.35$\times$ and 1.97--3.42$\times$, respectively.
Thus, architecture-specific state handling changes the efficiency
trade-off while preserving system gains on long histories.
The shifted crossover also motivates model-specific history-length and
load thresholds for path selection.

\subsection{Task Accuracy}

\begin{figure*}[!tbp]
    \centering
    \includegraphics[width=\textwidth,trim=0 1.25bp 0 6.75bp,clip]{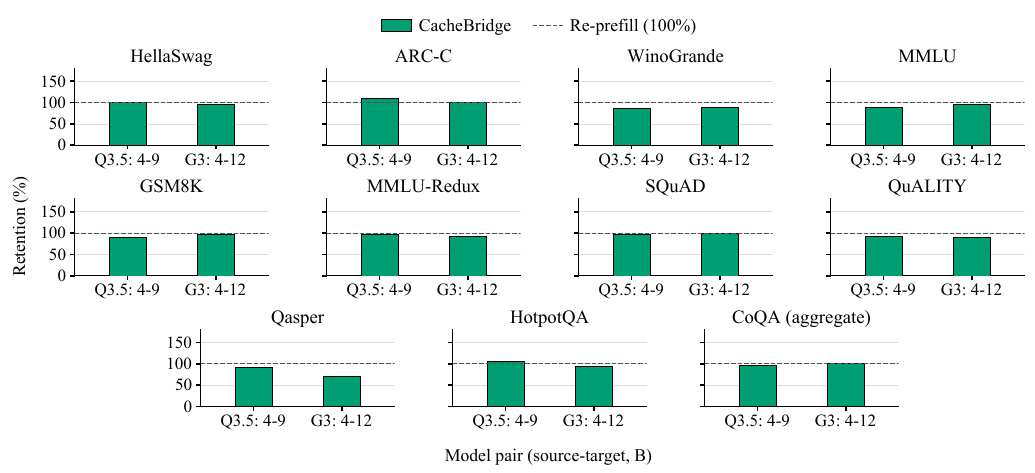}
    \caption{Task quality retention for the mixed-architecture extensions.
    Q3.5 and G3 denote Qwen3.5 and Gemma 3. CoQA reports one aggregate per pair.}
    \label{fig:accuracy-architecture}
\end{figure*}

For each model pair, the same architecture-adapted mapper
is evaluated on all 11 tasks. Calibration and evaluation data follow the
main method; the changes concern the architecture-specific state and
fitting steps above.
Figure~\ref{fig:accuracy-architecture} reports retention relative to the
matched re-prefill target.
Table~\ref{tab:accuracy-architecture} provides the complete numerical results.
CoQA reports one aggregate per pair.

\paragraph{Results.}
For Qwen3.5 4B$\rightarrow$9B, \method{} reaches at least re-prefill in three
of 11 settings, including 111.11\% on ARC-C and 104.88\% on HotpotQA.
Its lowest retention is 86.96\% on WinoGrande.
For Gemma 3 4B$\rightarrow$12B, retention ranges from 71.21\% on Qasper
to 100\% on ARC-C, and reaches 99.83\% on CoQA.
These measurements extend the evaluated model coverage beyond the three
families in the main accuracy experiment and show that the fitting
workflow can be reused after changing its state targets and structural
constraints.
\begin{table*}[!tbp]
\centering
\begingroup
\fontsize{8.5}{10.2}\selectfont
\setlength{\tabcolsep}{2pt}
\renewcommand{\arraystretch}{1.0}
\caption{Task retention (\%) for architecture-adapted configurations. Re-prefill is normalized to 100\%. CoQA reports aggregate retention.}\label{tab:accuracy-architecture}\label{tab:accuracy-complete-architecture}
\begin{tabular}{@{}p{0.100000\textwidth}>{\raggedleft\arraybackslash}p{\dimexpr0.081818\textwidth-2\tabcolsep\relax}>{\raggedleft\arraybackslash}p{\dimexpr0.081818\textwidth-2\tabcolsep\relax}>{\raggedleft\arraybackslash}p{\dimexpr0.081818\textwidth-2\tabcolsep\relax}>{\raggedleft\arraybackslash}p{\dimexpr0.081818\textwidth-2\tabcolsep\relax}>{\raggedleft\arraybackslash}p{\dimexpr0.081818\textwidth-2\tabcolsep\relax}>{\raggedleft\arraybackslash}p{\dimexpr0.081818\textwidth-2\tabcolsep\relax}>{\raggedleft\arraybackslash}p{\dimexpr0.081818\textwidth-2\tabcolsep\relax}>{\raggedleft\arraybackslash}p{\dimexpr0.081818\textwidth-2\tabcolsep\relax}>{\raggedleft\arraybackslash}p{\dimexpr0.081818\textwidth-2\tabcolsep\relax}>{\raggedleft\arraybackslash}p{\dimexpr0.081818\textwidth-2\tabcolsep\relax}>{\raggedleft\arraybackslash}p{\dimexpr0.081818\textwidth-2\tabcolsep\relax}@{}}
\toprule
Method & \shortstack{Hella\\Swag} & ARC-C & \shortstack{Wino\\Grande} & MMLU & GSM8K & \shortstack{MMLU-\\Redux} & SQuAD & QuALITY & Qasper & \shortstack{Hotpot\\QA} & CoQA \\
\midrule
\multicolumn{12}{@{}l}{\textbf{Qwen3.5 4B$\rightarrow$9B}} \\*
re-prefill & 100.00 & 100.00 & 100.00 & 100.00 & 100.00 & 100.00 & 100.00 & 100.00 & 100.00 & 100.00 & 100.00 \\*
\method{} & 100.00 & 111.11 & 86.96 & 88.46 & 89.29 & 96.67 & 98.12 & 92.86 & 91.56 & 104.88 & 96.96 \\
\midrule
\multicolumn{12}{@{}l}{\textbf{Gemma 3 4B$\rightarrow$12B}} \\*
re-prefill & 100.00 & 100.00 & 100.00 & 100.00 & 100.00 & 100.00 & 100.00 & 100.00 & 100.00 & 100.00 & 100.00 \\*
\method{} & 96.30 & 100.00 & 88.26 & 95.24 & 96.30 & 91.54 & 98.85 & 90.00 & 71.21 & 94.65 & 99.83 \\
\bottomrule
\end{tabular}
\endgroup
\end{table*}

\section{Mapper Fitting Diagnostics}
  \label{sec:appendix-fitting-diagnostics}

  We compare \method{} with the original CFM ridge mapper (Ridge) on
  HellaSwag and QuALITY using the diagnostic metrics of
  Heo et al.~\citep{heo2026crossmodel}. Table~\ref{tab:mapper-fitting-diagnostics}
  reports K/V reconstruction $R^2$, task retention, K/V error concentration
  and attention-output cosine similarity. All differences are
  $\Delta=\text{CacheBridge}-\text{Ridge}$. Retention is normalized to the
  matched re-prefill target, with its differences in percentage points
  (pp). The remaining differences use the original metric scales.
  \par
  \paragraph{Attention-output fidelity.}
  We prioritize preserving the target's attention output over maximizing
  coordinate-level Ridge fit quality. Ridge fits K/V coordinates with a
  regularized squared-error objective, while the target consumes the
  reconstructed state through the query-dependent attention output $O=AV$.
  Here $A$ contains attention weights over cached keys, and each value error
  contributes to the readout according to its attention weight.
  Attention-output fidelity thus captures how reconstruction affects the
  state passed to subsequent target computation. This principle motivates
  the value-map calibration with fixed key maps in
  Section~\ref{sec:fitting-readout}, which aligns reconstructed and
  re-prefill attention readouts at shared query states.
  \par
  On QuALITY, \method{} raises attention-output cosine over Ridge for all
  five model pairs, by $0.075$ to $0.330$. Qwen3 8B$\rightarrow$14B has a
  $2.973$ decrease in $R_V^2$ alongside a $0.195$ increase in
  attention-output cosine and a $52.82$\,pp retention gain. On QuALITY,
  attention-output fidelity thus reveals useful reconstruction properties
  beyond coordinate fit.
  \par
\begin{figure}[!tbp]
  \centering
  \includegraphics[width=\columnwidth,trim=0 219bp 0 1.75bp,clip]{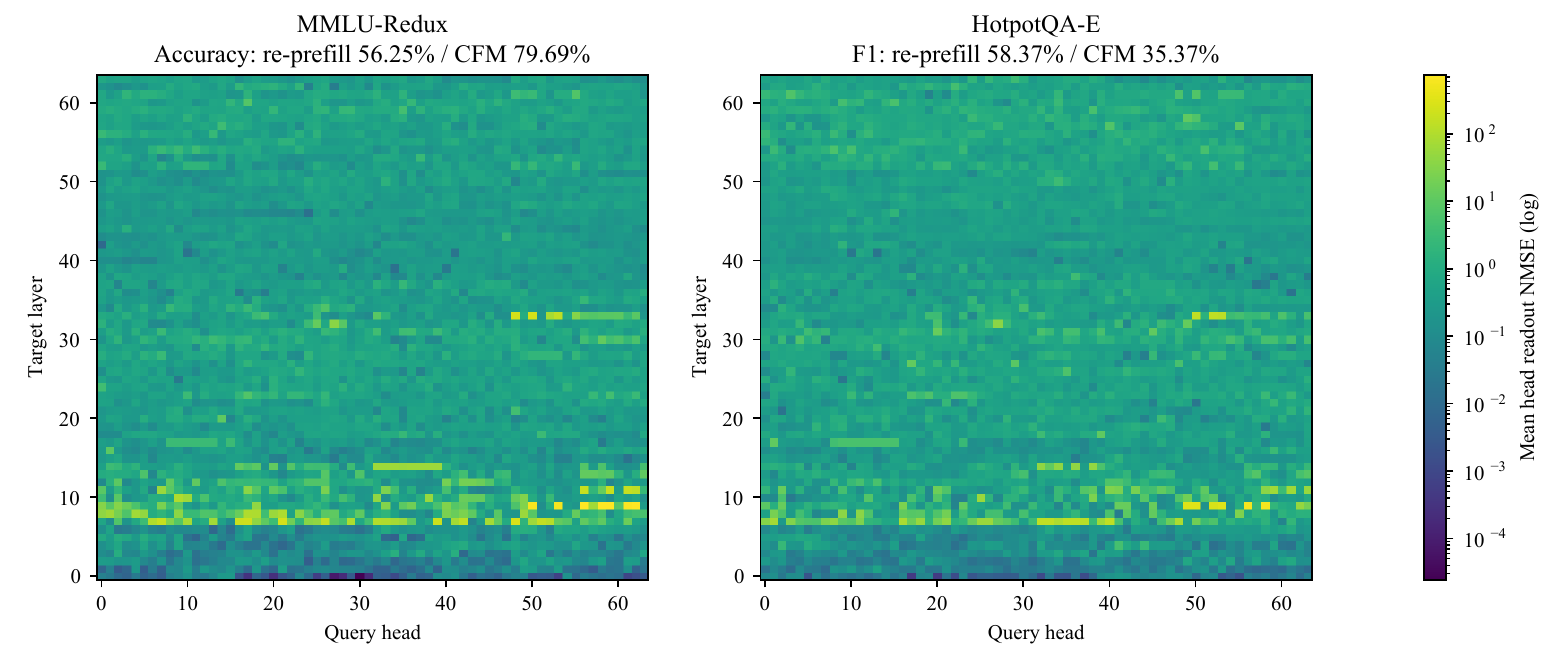}
  \caption{Full-Head CFM attention-readout NMSE.}
  \label{fig:long-context-cfm-readout}
  \end{figure}

  Figure~\ref{fig:long-context-cfm-readout} shows the Full-Head CFM
  attention-readout NMSE by target layer and query head for the
  long-context comparison in Section~\ref{sec:motivation-quality}.

\begin{table*}[!tbp]
\centering
\begingroup
\small
\setlength{\tabcolsep}{3pt}
\renewcommand{\arraystretch}{1.0}
\caption{Fitting diagnostics on HellaSwag and QuALITY.
Ridge denotes the original CFM ridge mapper, and
$\Delta=\text{CacheBridge}-\text{Ridge}$.
K-conc and V-conc denote K/V error concentration, and cosine denotes
attention-output cosine similarity.}
\label{tab:mapper-fitting-diagnostics}
\begin{tabular*}{\textwidth}{@{\extracolsep{\fill}}lrrrrrrrrr@{}}
\toprule
Pair & \shortstack{Ridge\\$R_K^2$} & $\Delta R_K^2$
& \shortstack{Ridge\\$R_V^2$} & $\Delta R_V^2$
& \shortstack{Ridge ret.\\(\%)} & \shortstack{$\Delta$ ret.\\(pp)}
& \shortstack{$\Delta$\\K-conc} & \shortstack{$\Delta$\\V-conc}
& \shortstack{$\Delta$\\cosine} \\
\midrule
\multicolumn{10}{@{}l}{\textbf{HellaSwag}} \\
Qwen3 14B$\to$32B & 0.674 & $-0.066$ & 0.575 & $-1.710$ & 98.75 & $+5.25$ & $-0.034$ & $-0.049$ & $-0.110$ \\
Qwen3 8B$\to$14B & 0.562 & $+0.002$ & 0.500 & $-3.527$ & 70.23 & $+25.77$ & $-0.272$ & $-0.266$ & $-0.051$ \\
Llama 3.1 8B$\to$70B & 0.727 & $-0.010$ & 0.438 & $-0.127$ & 100.00 & $0.00$ & $-0.078$ & $+0.001$ & $-0.018$ \\
Ministral 3 3B$\to$14B & 0.681 & $-0.015$ & 0.002 & $+0.445$ & 67.62 & $+26.42$ & $-0.103$ & $-0.014$ & $+0.237$ \\
Ministral 3 8B$\to$14B & 0.758 & $-0.030$ & $-0.518$ & $+1.086$ & 57.51 & $+44.49$ & $-0.147$ & $+0.023$ & $+0.499$ \\
\midrule
\multicolumn{10}{@{}l}{\textbf{QuALITY}} \\
Qwen3 14B$\to$32B & 0.662 & $-0.068$ & 0.564 & $-1.061$ & 90.13 & $+13.57$ & $-0.037$ & $-0.029$ & $+0.102$ \\
Qwen3 8B$\to$14B & 0.617 & $-0.061$ & 0.503 & $-2.973$ & 44.27 & $+52.82$ & $-0.093$ & $-0.300$ & $+0.195$ \\
Llama 3.1 8B$\to$70B & 0.653 & $-0.012$ & 0.336 & $-0.096$ & 77.66 & $+18.83$ & $-0.145$ & $-0.001$ & $+0.075$ \\
Ministral 3 3B$\to$14B & 0.650 & $-0.014$ & $-0.038$ & $+0.451$ & 48.37 & $+53.66$ & $-0.131$ & $-0.010$ & $+0.190$ \\
Ministral 3 8B$\to$14B & 0.739 & $-0.032$ & $-0.490$ & $+1.045$ & 35.61 & $+71.37$ & $-0.126$ & $+0.019$ & $+0.330$ \\
\bottomrule
\end{tabular*}
\endgroup
\end{table*}

\section{Complete \texorpdfstring{$\tau$}{tau}-bench Results}
  \label{sec:appendix-tau-bench}

  Table~\ref{tab:tau-bench-serving-complete} reports mean and P99
  end-to-end serving TTFT
  for Qwen3 14B$\rightarrow$32B retail-agent serving at every evaluated
  arrival rate. For each random seed, we construct a request trace by
  sampling uniformly with replacement from the $\tau$-bench retail
  workload. A Poisson arrival process at the configured QPS controls
  request arrivals, and each arrival rate pools 1,024 requests from two
  seeds.
  Table~\ref{tab:tau-bench-throughput-goodput} reports throughput and goodput.
  For each method and arrival rate, throughput divides completed requests
  by total run duration. Goodput uses the same duration and counts only
  requests satisfying both TTFT $\leq 1$\,s and complete-output latency
  $\leq 5$\,s.
  Table~\ref{tab:tau-bench-history-lengths} reports source-history lengths
  at handoff. The 5 and 24 QPS traces have similar lengths, with means
  of 4,779.72 and 4,775.02 tokens and the same reported quantiles.

\begin{table}[!tbp]
\centering
\begingroup
\fontsize{8.5}{10.2}\selectfont
\setlength{\tabcolsep}{3pt}
\renewcommand{\arraystretch}{1.0}
\caption{Complete $\tau$-bench serving TTFT for Qwen3 14B$\rightarrow$32B. All TTFT values are in seconds; each arrival rate pools 1,024 requests from two random seeds. Mean speedup is re-prefill mean TTFT divided by \method{} mean TTFT.}
\label{tab:tau-bench-serving-complete}
\begin{tabular}{@{}rrrrrr@{}}
\toprule
& \multicolumn{2}{c}{Re-prefill} & \multicolumn{2}{c}{\method{}} & \\
\cmidrule(lr){2-3}\cmidrule(lr){4-5}
QPS & Mean & P99 & Mean & P99 & \shortstack{Mean\\speedup} \\
\midrule
1 & 0.332 & 0.750 & 0.231 & 0.534 & 1.44$\times$ \\
2 & 0.465 & 1.702 & 0.273 & 0.738 & 1.70$\times$ \\
5 & 32.089 & 64.292 & 0.606 & 1.717 & 52.95$\times$ \\
8 & 50.013 & 99.171 & 10.219 & 24.833 & 4.89$\times$ \\
10 & 55.975 & 110.957 & 16.154 & 37.175 & 3.47$\times$ \\
12 & 60.021 & 118.476 & 19.598 & 40.251 & 3.06$\times$ \\
16 & 65.006 & 128.427 & 25.034 & 53.200 & 2.60$\times$ \\
24 & 70.112 & 138.374 & 30.033 & 62.648 & 2.33$\times$ \\
\bottomrule
\end{tabular}
\endgroup
\end{table}

\begin{table}[!tbp]
\centering
\begingroup
\fontsize{8.5}{10.2}\selectfont
\setlength{\tabcolsep}{3pt}
\renewcommand{\arraystretch}{1.0}
\caption{Complete $\tau$-bench throughput and goodput (requests/s). Goodput requires both TTFT $\leq 1$\,s and complete-output latency $\leq 5$\,s. Each method uses the same total run duration for its throughput and goodput.}
\label{tab:tau-bench-throughput-goodput}
\begin{tabular}{@{}rrrrr@{}}
\toprule
& \multicolumn{2}{c}{Throughput} & \multicolumn{2}{c}{Goodput} \\
\cmidrule(lr){2-3}\cmidrule(lr){4-5}
Arrival QPS & Re-prefill & \method{} & Re-prefill & \method{} \\
\midrule
1 & 1.042093 & 1.042249 & 1.042093 & 1.042249 \\
2 & 2.075756 & 2.077094 & 1.864937 & 2.077094 \\
5 & 3.172709 & 5.113431 & 0.024787 & 4.369387 \\
8 & 3.178250 & 5.988613 & 0.012415 & 0.228082 \\
10 & 3.179340 & 5.977268 & 0.009314 & 0.180952 \\
12 & 3.179768 & 6.180621 & 0.018631 & 0.162966 \\
16 & 3.180534 & 6.078236 & 0.012424 & 0.106844 \\
24 & 3.179121 & 6.089650 & 0.012418 & 0.095151 \\
\bottomrule
\end{tabular}
\endgroup
\end{table}

\paragraph{Prefill savings at 5 QPS.}
Table~\ref{tab:tau-bench-qps5-prefill} traces the 53.0$\times$ speedup
to the reduction in target-side prefill work. \method{} reduces actual
prefill tokens from 5,445,048 to 550,615, saving 4,894,433 tokens
(89.89\%) and cutting prefill work by 9.89$\times$.
Average prefill tokens per request fall from 5,317.43 to 537.71.
By reusing the source's processed history, \method{} reduces target
prefill demand and relieves queueing pressure at this arrival rate.
The resulting serving gain reaches 52.95$\times$ in mean TTFT,
from 32.089\,s to 0.606\,s.

\begin{table}[!tbp]
\centering
\begingroup
\fontsize{8.5}{10.2}\selectfont
\setlength{\tabcolsep}{3pt}
\renewcommand{\arraystretch}{1.0}
\caption{Prefill work and mean TTFT at 5 QPS on $\tau$-bench for
Qwen3 14B$\rightarrow$32B.}
\label{tab:tau-bench-qps5-prefill}
\begin{tabular}{@{}lrr@{}}
\toprule
Metric & Re-prefill & \method{} \\
\midrule
Actual prefill tokens, total & 5,445,048 & 550,615 \\
Mean prefill tokens per request & 5,317.43 & 537.71 \\
Mean TTFT (s) & 32.089 & 0.606 \\
\bottomrule
\end{tabular}
\endgroup
\end{table}

\begin{table*}[!tbp]
\centering
\begingroup
\fontsize{8.5}{10.2}\selectfont
\setlength{\tabcolsep}{8pt}
\renewcommand{\arraystretch}{1.0}
\caption{Source-history length $H$ at handoff in the $\tau$-bench serving workload (tokens).}
\label{tab:tau-bench-history-lengths}
\begin{tabular}{@{}rrrrrrrr@{}}
\toprule
QPS & Min & Mean & P50 & P90 & P95 & P99 & Max \\
\midrule
5 & 4,356 & 4,779.72 & 4,753 & 4,987 & 5,221 & 5,298 & 5,298 \\
24 & 4,356 & 4,775.02 & 4,753 & 4,987 & 5,221 & 5,298 & 5,298 \\
\bottomrule
\end{tabular}
\endgroup
\end{table*}

\section{Serving Efficiency Analysis}
\label{sec:appendix-efficiency-analysis}
\label{sec:appendix-cfm-serving}

We compare \method{}, CFM and re-prefill on Qwen3 14B$\rightarrow$32B
at 5 and 24 QPS to separate handoff service time from the waiting that
accumulates before service.
Table~\ref{tab:serving-efficiency-stages} reports mean durations from
front-end waiting and admission through client-visible output.
The complete-handoff column measures cache preparation and installation
before engine submission. End-to-end serving TTFT also includes waiting,
submission and target execution through first-token delivery.
These stage profiles describe the efficiency-analysis runs.
Appendix~\ref{sec:appendix-tau-bench} reports the serving results over
1,024 pooled requests per arrival rate.

\begin{table*}[!tbp]
\centering
\begingroup
\fontsize{8.5}{10.2}\selectfont
\setlength{\tabcolsep}{3pt}
\renewcommand{\arraystretch}{1.0}
\caption{Mean stage durations in the efficiency-analysis runs (seconds). Stage sum adds the displayed rounded durations.}
\label{tab:serving-efficiency-stages}
\begin{tabular*}{\textwidth}{@{\extracolsep{\fill}}rlrrrrrr@{}}
\toprule
QPS & Method & \shortstack{Front-end wait\\/ admission} & \shortstack{Complete\\handoff} & \shortstack{Submission\\preparation} & \shortstack{Engine submit\\to first token} & \shortstack{First token\\to client} & \shortstack{Stage\\sum} \\
\midrule
5 & \method{} & \textbf{0.308} & \textbf{0.112} & 0.007 & 0.097 & 0.082 & 0.606 \\
5 & CFM & 166.006 & 0.792 & 0.007 & 11.464 & 0.713 & 178.982 \\
5 & Re-prefill & 25.636 & 0.000 & 0.010 & 6.431 & 0.012 & 32.089 \\
\midrule
24 & \method{} & \textbf{28.205} & \textbf{0.122} & 0.006 & 1.232 & 0.105 & 29.670 \\
24 & CFM & 203.177 & 0.791 & 0.007 & 11.111 & 0.680 & 215.766 \\
24 & Re-prefill & 63.136 & 0.000 & 0.009 & 6.730 & 0.010 & 69.885 \\
\bottomrule
\end{tabular*}
\endgroup
\end{table*}

\paragraph{Queue amplification of handoff savings.}
At 5 QPS, \method{} reduces the handoff service interval from 0.792\,s with
CFM to 0.112\,s, saving 0.680\,s per handoff.
At 24 QPS, the corresponding reduction is 0.669\,s, from 0.791\,s to
0.122\,s. This roughly 0.68\,s saving has a much larger effect on
queued requests when handoffs serialize: every waiting request inherits
the handoff service demand ahead of it.
Front-end waiting and admission fall from 166.006\,s to 0.308\,s at
5 QPS, and from 203.177\,s to 28.205\,s at 24 QPS.
Thus, the serving gain reflects both a faster handoff and the reduced
backlog of handoffs ahead of each request.
\par
Relative to re-prefill, transfer also removes repeated history prefill
from the target. At 5 QPS, the engine-submission-to-first-token interval
falls from 6.431\,s to 0.097\,s, while front-end waiting and admission
fall from 25.636\,s to 0.308\,s.
Together, the displayed stages sum to 32.089\,s for re-prefill and
0.606\,s for \method{}, matching the 53.0$\times$ mean-TTFT result.
At 5 QPS, the 53.0$\times$ mean-TTFT
gain on these finite request traces primarily reflects reduced queueing:
waiting before admission or handoff falls from 25.64 to 0.31\,s,
explaining 80.5\% of the absolute mean-TTFT reduction
(Table~\ref{tab:serving-efficiency-stages}).

The roughly 0.68\,s handoff saving is the comparison with CFM, while
the 53.0$\times$ result uses re-prefill as its reference.

The following tables provide the detailed request and handoff profiles for CFM.

\paragraph{Detailed CFM trace at 24 QPS.}
A separate 512-request diagnostic CFM trace, built with the same sampling
and arrival rules as the main experiment
(Appendix~\ref{sec:appendix-tau-bench}), provides a finer breakdown, with
mean request TTFT of 215.767115\,s. It also supplies the 24 QPS CFM row of
Table~\ref{tab:serving-efficiency-stages}. Table~\ref{tab:cfm-qps24-request} separates
queueing, handoff and engine intervals.
Waiting for controller acceptance and resource admission accounts for
94.16\% of mean TTFT. Client-side arrival generation and local delivery
together average only 0.120\,ms, locating the backlog primarily in the
serving front end and resource admission queues.

\begin{table*}[!tbp]
\centering
\begingroup
\fontsize{8.5}{10.2}\selectfont
\setlength{\tabcolsep}{10pt}
\renewcommand{\arraystretch}{1.0}
\caption{CFM request TTFT breakdown at 24 QPS. Shares use mean request TTFT as the denominator.}
\label{tab:cfm-qps24-request}
\begin{tabular}{@{}lrr@{}}
\toprule
Request stage & Mean time (s) & TTFT share (\%) \\
\midrule
Ingress receipt to controller acceptance & 142.317770 & 65.9590 \\
Controller acceptance to resource admission & 60.856028 & 28.2045 \\
Other arrival and dispatch intervals & 0.003297 & 0.0015 \\
Complete CFM handoff & 0.790838 & 0.3665 \\
Prefix publication to engine submission & 0.007358 & 0.0034 \\
Engine submission to first streamed token & 11.111444 & 5.1497 \\
First streamed token to client visibility & 0.680380 & 0.3153 \\
\midrule
Total & 215.767115 & 100 \\
\bottomrule
\end{tabular}
\endgroup
\end{table*}

\paragraph{Handoff composition.}
The complete CFM handoff averages 790.838\,ms
(Table~\ref{tab:cfm-qps24-handoff}). Target apply RPC accounts for
91.47\% of this interval and includes feature delivery, FP32 layout and
concatenation, mapping, RoPE, installation, TP coordination and
synchronization waits. Its share measures the complete RPC interval,
with matrix multiplication and mapper kernels forming only part of it.

\begin{table*}[!tbp]
\centering
\begingroup
\fontsize{8.5}{10.2}\selectfont
\setlength{\tabcolsep}{10pt}
\renewcommand{\arraystretch}{1.0}
\caption{CFM handoff breakdown at 24 QPS. Shares use the 790.838\,ms mean handoff as the denominator.}
\label{tab:cfm-qps24-handoff}
\begin{tabular}{@{}lrr@{}}
\toprule
Handoff stage & Mean time (ms) & Handoff share (\%) \\
\midrule
Source export RPC & 28.836 & 3.646 \\
Target apply RPC & 723.342 & 91.465 \\
Source release RPC & 2.508 & 0.317 \\
Remaining host protocol and bridging intervals & 36.152 & 4.571 \\
\bottomrule
\end{tabular}
\endgroup
\end{table*}

\paragraph{Per-request share and run-window coverage.}
The handoff occupies only 0.37\% of mean request TTFT, yet the 512
handoffs execute without overlap and sum to 404.909\,s, covering 93.56\%
of the 432.776\,s window ending at the last complete output.
Target apply RPC alone sums to 370.351\,s, covering 85.58\% of the same
window. Thus, the small per-request handoff share reflects the large
queueing denominator, while serialized handoffs occupy most of the run.

All requests are scheduled to arrive within 20.942\,s, followed by
411.834\,s of draining. The finite-trace completion rate is
$512/432.776=1.183$ requests/s. Mean TTFT grows from 29.773\,s for the
first 64 requests to 395.275\,s for the last 64, while mean handoff time
in each 64-request group remains within 0.786--0.797\,s.
The growing TTFT therefore tracks accumulated admission waiting as the
serialized handoff path drains the offered load.
\par

\section{Large-to-Small Handoffs}
\label{sec:appendix-large-to-small}

We evaluate reverse-direction transfer on Qwen3 32B$\rightarrow$14B,
Qwen3 14B$\rightarrow$8B, Llama 3.1 70B$\rightarrow$8B,
Ministral 3 14B$\rightarrow$3B, and Ministral 3 14B$\rightarrow$8B.
Each pair covers six chat and reasoning tasks, four context-grounded tasks,
and one CoQA aggregate, for 55 model-pair/task settings.
As in Section~\ref{sec:eval-accuracy}, we report retention rate relative to
each setting's matched re-prefill target, with re-prefill normalized to 100\%.
\begin{figure*}[!tbp]
    \centering
    \includegraphics[width=\textwidth,trim=0 2bp 0 7bp,clip]{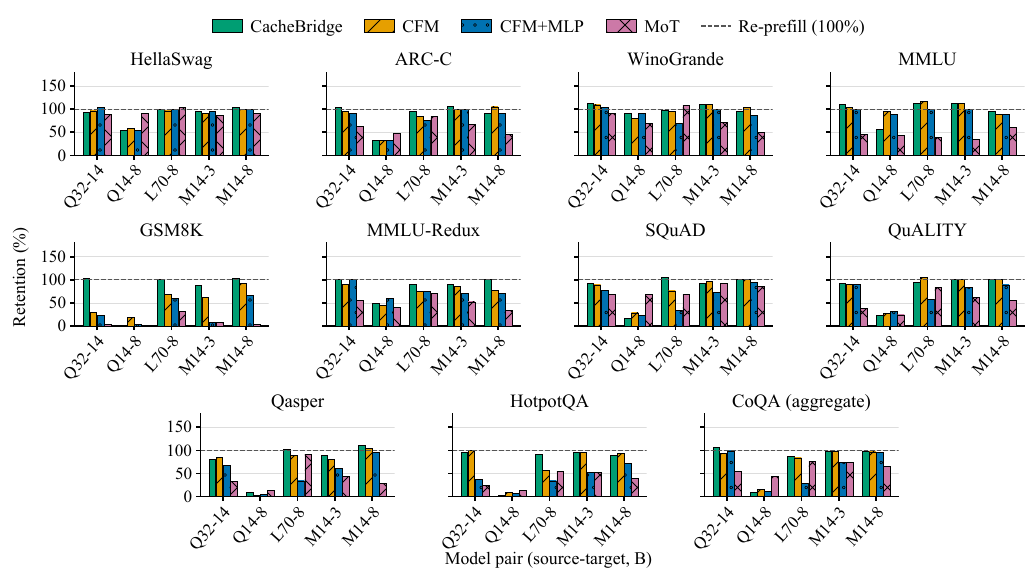}
    \caption{Large-to-small task quality retention relative to re-prefill.
    Q, L and M denote Qwen3, Llama 3.1 and Ministral 3. CoQA reports one aggregate per pair.}
    \label{fig:accuracy-reverse-chat}
    \label{fig:accuracy-reverse-context}
\end{figure*}

\paragraph{Chat and reasoning.}
Figure~\ref{fig:accuracy-reverse-chat} includes the six reasoning tasks.
\method{} matches or exceeds the strongest transfer baseline in 19 of
30 settings, including five ties, and reaches at least re-prefill in 14.
For Qwen3 32B$\rightarrow$14B, retention ranges from 92.31\% to 113.64\%
on these tasks. Qwen3 14B$\rightarrow$8B exhibits a larger quality gap,
with retention of 0\% on GSM8K and 33.33\% on ARC-C.

\paragraph{Context-grounded tasks and conversations.}
Across the 25 context-grounded and CoQA settings in
Figure~\ref{fig:accuracy-reverse-context},
\method{} matches or exceeds the strongest transfer baseline in 13,
including two ties.
Across all 55 reverse-direction settings, the corresponding count is 32,
including seven ties.
Over the 11 tasks, retention ranges from 81.46\% to 113.64\% for
Qwen3 32B$\rightarrow$14B, 87.12\% to 113.04\% for Llama 3.1
70B$\rightarrow$8B, 88.46\% to 113.04\% for Ministral 3
14B$\rightarrow$3B, and 88.50\% to 110.64\% for Ministral 3
14B$\rightarrow$8B.
Qwen3 14B$\rightarrow$8B instead ranges from 0\% to 90\%, with
2.28\% on HotpotQA and 9.37\% on CoQA.
Four of the five reverse pairs thus retain 81.46--113.64\% of re-prefill
quality on all 11 tasks, with each comparison normalized to the smaller
target.
Table~\ref{tab:accuracy-complete-reverse} provides the complete results.
\begin{table*}[!tbp]
\centering
\begingroup
\fontsize{8.5}{10.2}\selectfont
\setlength{\tabcolsep}{2pt}
\renewcommand{\arraystretch}{1.0}
\caption{Complete large-to-small retention (\%). Re-prefill is normalized to 100\%. CoQA reports aggregate retention. Bold marks the highest transfer retention, including ties.}\label{tab:accuracy-complete-reverse}
\begin{tabular}{@{}p{0.100000\textwidth}>{\raggedleft\arraybackslash}p{\dimexpr0.081818\textwidth-2\tabcolsep\relax}>{\raggedleft\arraybackslash}p{\dimexpr0.081818\textwidth-2\tabcolsep\relax}>{\raggedleft\arraybackslash}p{\dimexpr0.081818\textwidth-2\tabcolsep\relax}>{\raggedleft\arraybackslash}p{\dimexpr0.081818\textwidth-2\tabcolsep\relax}>{\raggedleft\arraybackslash}p{\dimexpr0.081818\textwidth-2\tabcolsep\relax}>{\raggedleft\arraybackslash}p{\dimexpr0.081818\textwidth-2\tabcolsep\relax}>{\raggedleft\arraybackslash}p{\dimexpr0.081818\textwidth-2\tabcolsep\relax}>{\raggedleft\arraybackslash}p{\dimexpr0.081818\textwidth-2\tabcolsep\relax}>{\raggedleft\arraybackslash}p{\dimexpr0.081818\textwidth-2\tabcolsep\relax}>{\raggedleft\arraybackslash}p{\dimexpr0.081818\textwidth-2\tabcolsep\relax}>{\raggedleft\arraybackslash}p{\dimexpr0.081818\textwidth-2\tabcolsep\relax}@{}}
\toprule
Method & \shortstack{Hella\\Swag} & ARC-C & \shortstack{Wino\\Grande} & MMLU & GSM8K & \shortstack{MMLU-\\Redux} & SQuAD & QuALITY & Qasper & \shortstack{Hotpot\\QA} & CoQA \\
\midrule
\multicolumn{12}{@{}l}{\textbf{Qwen3 32B$\rightarrow$14B}} \\*
re-prefill & 100.00 & 100.00 & 100.00 & 100.00 & 100.00 & 100.00 & 100.00 & 100.00 & 100.00 & 100.00 & 100.00 \\*
CFM & 96.15 & 95.24 & 109.09 & 103.85 & 29.63 & 90.00 & 88.95 & 89.66 & \textbf{85.42} & \textbf{100.28} & 93.72 \\*
MLP & \textbf{103.85} & 90.48 & 104.55 & 100.00 & 22.22 & \textbf{100.00} & 77.24 & 89.66 & 66.94 & 36.69 & 99.02 \\*
MoT & 88.46 & 61.90 & 90.91 & 46.15 & 3.70 & 55.00 & 68.12 & 37.93 & 32.54 & 24.73 & 55.09 \\*
\method{} & 92.31 & \textbf{104.76} & \textbf{113.64} & \textbf{111.54} & \textbf{103.70} & \textbf{100.00} & \textbf{92.74} & \textbf{93.10} & 81.46 & 95.28 & \textbf{106.69} \\
\midrule
\multicolumn{12}{@{}l}{\textbf{Qwen3 14B$\rightarrow$8B}} \\*
re-prefill & 100.00 & 100.00 & 100.00 & 100.00 & 100.00 & 100.00 & 100.00 & 100.00 & 100.00 & 100.00 & 100.00 \\*
CFM & 58.33 & 33.33 & 80.00 & \textbf{96.00} & \textbf{17.86} & 45.00 & 28.23 & 28.00 & 2.97 & 9.49 & 15.36 \\*
MLP & 54.17 & 33.33 & \textbf{90.00} & 88.00 & 3.57 & \textbf{60.00} & 22.14 & \textbf{32.00} & 4.23 & 6.38 & 11.31 \\*
MoT & \textbf{91.67} & \textbf{47.62} & 70.00 & 44.00 & 0.00 & 40.00 & \textbf{68.78} & 24.00 & \textbf{14.14} & \textbf{13.55} & \textbf{43.59} \\*
\method{} & 54.17 & 33.33 & \textbf{90.00} & 56.00 & 0.00 & 50.00 & 16.31 & 24.00 & 9.37 & 2.28 & 9.37 \\
\midrule
\multicolumn{12}{@{}l}{\textbf{Llama 3.1 70B$\rightarrow$8B}} \\*
re-prefill & 100.00 & 100.00 & 100.00 & 100.00 & 100.00 & 100.00 & 100.00 & 100.00 & 100.00 & 100.00 & 100.00 \\*
CFM & 96.15 & 85.00 & 95.65 & \textbf{117.39} & 69.23 & 75.00 & 76.07 & \textbf{105.26} & 89.65 & 57.35 & 83.84 \\*
MLP & 100.00 & 75.00 & 69.57 & 100.00 & 60.00 & 75.00 & 33.95 & 57.89 & 33.88 & 34.11 & 29.19 \\*
MoT & \textbf{103.85} & 85.00 & \textbf{108.70} & 39.13 & 30.77 & 70.00 & 68.93 & 84.21 & 91.31 & 54.03 & 75.84 \\*
\method{} & 100.00 & \textbf{95.00} & 96.96 & 113.04 & \textbf{100.00} & \textbf{90.00} & \textbf{104.41} & 94.21 & \textbf{102.20} & \textbf{90.62} & \textbf{87.12} \\
\midrule
\multicolumn{12}{@{}l}{\textbf{Ministral 3 14B$\rightarrow$3B}} \\*
re-prefill & 100.00 & 100.00 & 100.00 & 100.00 & 100.00 & 100.00 & 100.00 & 100.00 & 100.00 & 100.00 & 100.00 \\*
CFM & 91.30 & 100.00 & \textbf{109.52} & \textbf{113.04} & 61.54 & 85.71 & \textbf{96.73} & \textbf{100.00} & 80.40 & \textbf{96.53} & 97.97 \\*
MLP & \textbf{95.65} & 100.00 & 100.00 & 100.00 & 7.69 & 71.43 & 71.96 & 83.33 & 61.56 & 53.19 & 74.30 \\*
MoT & 86.96 & 66.67 & 71.43 & 34.78 & 7.69 & 52.38 & 92.52 & 61.11 & 44.59 & 52.97 & 74.55 \\*
\method{} & \textbf{95.65} & \textbf{106.67} & \textbf{109.52} & \textbf{113.04} & \textbf{88.46} & \textbf{90.48} & 92.96 & \textbf{100.00} & \textbf{88.80} & 96.34 & \textbf{98.80} \\
\midrule
\multicolumn{12}{@{}l}{\textbf{Ministral 3 14B$\rightarrow$8B}} \\*
re-prefill & 100.00 & 100.00 & 100.00 & 100.00 & 100.00 & 100.00 & 100.00 & 100.00 & 100.00 & 100.00 & 100.00 \\*
CFM & 100.00 & \textbf{105.00} & \textbf{104.55} & 88.00 & 92.31 & 76.19 & \textbf{101.63} & \textbf{100.00} & 104.17 & \textbf{93.28} & 96.88 \\*
MLP & 100.00 & 90.00 & 86.36 & 88.00 & 65.38 & 71.43 & 94.74 & 88.89 & 95.82 & 71.17 & 95.51 \\*
MoT & 91.67 & 45.00 & 50.00 & 60.00 & 3.85 & 33.33 & 85.51 & 55.56 & 27.99 & 39.86 & 64.78 \\*
\method{} & \textbf{104.17} & 90.00 & 95.45 & \textbf{96.00} & \textbf{103.85} & \textbf{100.00} & 100.54 & \textbf{100.00} & \textbf{110.64} & 88.50 & \textbf{97.30} \\
\bottomrule
\end{tabular}
\endgroup
\end{table*}

\section{Recovering Multi-Turn Quality on Qwen3 8B\texorpdfstring{$\rightarrow$}{ to }14B}
\label{sec:appendix-qwen-failure}

We examine the CoQA quality loss for Qwen3 8B$\rightarrow$14B through
handoffs near the end of a conversation and an expanded mapper, which
raises retention averaged over handoff turns 1--10 from 31.07\% to 89.11\%.
These experiments characterize the quality gap across continuation lengths
and show its recovery with a wider reconstruction space.

\paragraph{Handoffs near conversation end.}
We place one handoff with 1, 2 or 5 turns remaining.
Table~\ref{tab:coqa-remaining-all} scores all post-handoff turns, while
Table~\ref{tab:coqa-remaining-final} scores only the final turn.
Each ratio uses its matched re-prefill reference.
The one-turn-remaining entries coincide because the two scoring scopes
cover the same remaining answer.
For Qwen3 8B$\rightarrow$14B, \method{} retains 17.69\%, 18.94\% and
25.30\% when scoring all 1, 2 and 5 remaining turns, respectively.
Thus, the quality gap persists even when only a single post-handoff
answer is evaluated.
For Ministral 3 8B$\rightarrow$14B, the corresponding retention rates
are 92.08\%, 96.02\% and 96.04\%, showing that the same handoff protocol
preserves substantially more quality on another model pair.

\paragraph{Expanded mapper.}
For Qwen3 8B$\rightarrow$14B, we augment the main method with a full
residual mapping over all 36 source layers and two KV heads per group.
This increases the mapper size from 83.97M to 859.11M parameters.
Averaged over handoff turns 1--10, CoQA retention rises from 31.07\% to
89.11\% (Section~\ref{sec:eval-accuracy}).
Table~\ref{tab:coqa-f41} reports the expanded mapper's retention
and matched re-prefill scores for the handoff protocols above.

\paragraph{Relation to the compact mapper.}
For this pair, the main mapper uses eight selected source layers and one
source KV head per selected layer for each target head
(Section~\ref{sec:fitting-interface} and
Table~\ref{tab:calibration-support}).
Value-map calibration adjusts the existing value-map weights and bias
within this fixed head assignment (Section~\ref{sec:fitting-readout}).
The expanded residual accesses a wider set of source-layer and head
features. The observed quality recovery shows that this model pair's
conversational continuation benefits from a broader reconstruction space.
Its 10.23$\times$ parameter count illustrates the tradeoff between
reconstruction capacity and the compact mapper deployed by the online
system (Section~\ref{sec:implementation}).

\begin{table*}[!tbp]
\centering
\small
\setlength{\tabcolsep}{6pt}
\caption{CoQA retention rates (\%) over all post-handoff turns.}
\label{tab:coqa-remaining-all}
\begin{tabular}{@{}llrrrr@{}}
\toprule
Model pair & Turns remaining & CFM & MLP & MoT & CacheBridge \\
\midrule
Qwen3 8B$\rightarrow$14B & 1 & \textbf{47.45} & 10.29 & 47.41 & 17.69 \\
 & 2 & 45.20 & 7.30 & \textbf{53.63} & 18.94 \\
 & 5 & 35.67 & 7.73 & \textbf{50.04} & 25.30 \\
\midrule
Qwen3 14B$\rightarrow$32B & 1 & \textbf{126.14} & 124.49 & 100.81 & 116.98 \\
 & 2 & 132.83 & \textbf{133.37} & 113.01 & 118.87 \\
 & 5 & \textbf{121.97} & 116.91 & 106.14 & 108.09 \\
\midrule
Llama 3.1 8B$\rightarrow$70B & 1 & 69.47 & 64.40 & 88.73 & \textbf{99.93} \\
 & 2 & 64.37 & 60.75 & \textbf{91.39} & 87.08 \\
 & 5 & 66.07 & 54.38 & 95.00 & \textbf{97.17} \\
\midrule
Ministral 3 3B$\rightarrow$14B & 1 & 23.46 & 47.83 & 64.33 & \textbf{84.19} \\
 & 2 & 24.17 & 50.09 & 68.60 & \textbf{90.33} \\
 & 5 & 18.04 & 38.91 & 69.19 & \textbf{88.97} \\
\midrule
Ministral 3 8B$\rightarrow$14B & 1 & 4.50 & 60.20 & 60.00 & \textbf{92.08} \\
 & 2 & 9.07 & 64.24 & 63.42 & \textbf{96.02} \\
 & 5 & 8.02 & 59.31 & 65.60 & \textbf{96.04} \\
\bottomrule
\end{tabular}
\end{table*}

\begin{table*}[!tbp]
\centering
\small
\setlength{\tabcolsep}{6pt}
\caption{CoQA retention rates (\%) on the final turn.}
\label{tab:coqa-remaining-final}
\begin{tabular}{@{}llrrrr@{}}
\toprule
Model pair & Turns remaining & CFM & MLP & MoT & CacheBridge \\
\midrule
Qwen3 8B$\rightarrow$14B & 1 & \textbf{47.45} & 10.29 & 47.41 & 17.69 \\
 & 2 & 44.42 & 5.32 & \textbf{54.58} & 15.15 \\
 & 5 & 34.84 & 2.12 & \textbf{45.90} & 18.52 \\
\midrule
Qwen3 14B$\rightarrow$32B & 1 & \textbf{126.14} & 124.49 & 100.81 & 116.98 \\
 & 2 & 121.34 & \textbf{122.79} & 104.88 & 101.75 \\
 & 5 & \textbf{124.77} & 118.41 & 107.96 & 103.04 \\
\midrule
Llama 3.1 8B$\rightarrow$70B & 1 & 69.47 & 64.40 & 88.73 & \textbf{99.93} \\
 & 2 & 60.66 & 57.84 & \textbf{95.44} & 86.13 \\
 & 5 & 56.32 & 47.04 & \textbf{93.76} & 87.97 \\
\midrule
Ministral 3 3B$\rightarrow$14B & 1 & 23.46 & 47.83 & 64.33 & \textbf{84.19} \\
 & 2 & 23.30 & 41.93 & 71.37 & \textbf{85.35} \\
 & 5 & 18.87 & 36.34 & 69.81 & \textbf{86.07} \\
\midrule
Ministral 3 8B$\rightarrow$14B & 1 & 4.50 & 60.20 & 60.00 & \textbf{92.08} \\
 & 2 & 8.75 & 54.13 & 58.31 & \textbf{94.12} \\
 & 5 & 5.75 & 49.85 & 68.31 & \textbf{94.30} \\
\bottomrule
\end{tabular}
\end{table*}

\begin{table}[!tbp]
\centering
\begingroup
\fontsize{8.5}{10.2}\selectfont
\setlength{\tabcolsep}{3pt}
\renewcommand{\arraystretch}{1.0}
\caption{CoQA results for the expanded mapper on Qwen3 8B$\rightarrow$14B. Re-prefill gives raw scores (0--100), and the expanded mapper gives retention (\%).}
\label{tab:coqa-f41}
\begin{tabular*}{\columnwidth}{@{\extracolsep{\fill}}lrr@{}}
\toprule
Turn & \shortstack[r]{re-prefill\\score} & \shortstack[r]{Expanded mapper\\(\%)} \\
\midrule
1 & 72.79 & 84.14 \\
2 & 63.92 & 107.70 \\
3 & 69.54 & 86.82 \\
4 & 61.34 & 93.58 \\
5 & 59.81 & 99.37 \\
6 & 69.20 & 95.85 \\
7 & 61.22 & 82.39 \\
8 & 69.63 & 79.10 \\
9 & 67.05 & 79.89 \\
10 & 70.49 & 82.27 \\
Mean (turns 1--10) & 66.50 & 89.11 \\
Last 1, all & 71.40 & 77.59 \\
Last 1, final & 71.40 & 77.59 \\
Last 2, all & 71.39 & 79.64 \\
Last 2, final & 74.28 & 84.71 \\
Last 5, all & 70.90 & 87.47 \\
Last 5, final & 73.47 & 85.98 \\
\bottomrule
\end{tabular*}
\endgroup
\end{table}

\newcommand{\CompleteAccuracyHeading}{%
\section{Complete Task Accuracy Results}
  \label{sec:appendix-complete-accuracy}

  The following tables report the main accuracy results for the five model
  pairs and the turn-indexed CoQA retention.
  Transfer retention uses each setting's matched re-prefill reference.
  Original scores and normalized references are identified in the captions,
  and the CoQA protocols retain their separate aggregation scopes.
\par
}

\begin{table*}[!tbp]
\CompleteAccuracyHeading
\centering
\begingroup
\fontsize{8.5}{10.2}\selectfont
\setlength{\tabcolsep}{2pt}
\renewcommand{\arraystretch}{1.0}
\caption{Complete chat, reasoning and RAG results. Re-prefill rows give raw scores (0--100); transfer rows give retention (\%). Bold marks the highest transfer retention for each task, including ties.}\label{tab:accuracy-complete-main}\label{tab:accuracy-chat}\label{tab:accuracy-rag}
\begin{tabular}{@{}p{0.120000\textwidth}>{\raggedleft\arraybackslash}p{\dimexpr0.088000\textwidth-2\tabcolsep\relax}>{\raggedleft\arraybackslash}p{\dimexpr0.088000\textwidth-2\tabcolsep\relax}>{\raggedleft\arraybackslash}p{\dimexpr0.088000\textwidth-2\tabcolsep\relax}>{\raggedleft\arraybackslash}p{\dimexpr0.088000\textwidth-2\tabcolsep\relax}>{\raggedleft\arraybackslash}p{\dimexpr0.088000\textwidth-2\tabcolsep\relax}>{\raggedleft\arraybackslash}p{\dimexpr0.088000\textwidth-2\tabcolsep\relax}>{\raggedleft\arraybackslash}p{\dimexpr0.088000\textwidth-2\tabcolsep\relax}>{\raggedleft\arraybackslash}p{\dimexpr0.088000\textwidth-2\tabcolsep\relax}>{\raggedleft\arraybackslash}p{\dimexpr0.088000\textwidth-2\tabcolsep\relax}>{\raggedleft\arraybackslash}p{\dimexpr0.088000\textwidth-2\tabcolsep\relax}@{}}
\toprule
Method & \shortstack{Hella\\Swag} & ARC-C & \shortstack{Wino\\Grande} & MMLU & GSM8K & \shortstack{MMLU-\\Redux} & \shortstack{SQuAD\\v1.1} & QuALITY & Qasper & \shortstack{Hotpot\\QA-E} \\
\midrule
\multicolumn{11}{@{}l}{\textbf{Qwen3 8B$\rightarrow$14B}} \\*
re-prefill & 78.12 & 61.18 & 74.98 & 83.33 & 96.89 & 68.02 & 21.78 & 80.49 & 39.35 & 66.06 \\*
CFM & 70.23 & 65.03 & 75.54 & 88.51 & 9.72 & 75.12 & 45.41 & 44.27 & 37.07 & 19.81 \\*
MLP & 42.00 & 45.00 & 52.08 & 63.16 & 1.11 & 75.18 & 22.47 & 30.21 & 2.87 & 4.01 \\*
MoT & \textbf{96.00} & 77.50 & 81.25 & 34.74 & 2.05 & 39.75 & 71.08 & 33.07 & 45.15 & 36.24 \\*
\method{} & \textbf{96.00} & \textbf{93.46} & \textbf{93.75} & \textbf{95.79} & \textbf{100.00} & \textbf{94.59} & \textbf{85.39} & \textbf{97.09} & \textbf{96.78} & \textbf{88.02} \\
\midrule
\multicolumn{11}{@{}l}{\textbf{Qwen3 14B$\rightarrow$32B}} \\*
re-prefill & 78.12 & 54.69 & 65.59 & 85.26 & 45.26 & 66.65 & 26.20 & 84.37 & 41.46 & 62.25 \\*
CFM & 98.75 & 104.65 & 100.00 & 95.47 & \textbf{164.49} & \textbf{113.96} & \textbf{104.59} & 90.13 & 75.03 & 66.57 \\*
MLP & 102.00 & \textbf{111.43} & 95.24 & 93.81 & 125.13 & 108.90 & 94.39 & 81.30 & 60.97 & 40.87 \\*
MoT & 98.00 & 100.00 & \textbf{114.29} & \textbf{101.03} & 137.35 & 102.80 & 99.89 & 89.35 & \textbf{96.53} & 96.47 \\*
\method{} & \textbf{104.00} & \textbf{111.43} & 111.90 & 97.54 & 131.99 & 101.04 & 97.36 & \textbf{103.70} & 93.34 & \textbf{96.64} \\
\midrule
\multicolumn{11}{@{}l}{\textbf{Llama 3.1 8B$\rightarrow$70B}} \\*
re-prefill & 81.25 & 60.92 & 81.22 & 83.33 & 80.14 & 77.22 & 82.87 & 83.70 & 16.20 & 15.77 \\*
CFM & \textbf{100.00} & 100.00 & 90.38 & 84.21 & 36.14 & 69.42 & 49.55 & 77.66 & 80.12 & 69.53 \\*
MLP & 96.15 & 89.74 & 86.54 & 64.21 & 3.88 & 61.81 & 19.60 & 52.47 & 16.09 & \textbf{113.04} \\*
MoT & \textbf{100.00} & \textbf{102.56} & \textbf{100.00} & 98.95 & \textbf{94.70} & \textbf{99.29} & \textbf{96.37} & 93.77 & 38.11 & 37.00 \\*
\method{} & \textbf{100.00} & 100.00 & \textbf{100.00} & \textbf{100.00} & 93.60 & 85.42 & 83.16 & \textbf{96.49} & \textbf{109.75} & 93.85 \\
\midrule
\multicolumn{11}{@{}l}{\textbf{Ministral 3 3B$\rightarrow$14B}} \\*
re-prefill & 77.20 & 65.61 & 78.14 & 77.19 & 87.34 & 76.62 & 89.23 & 73.25 & 28.67 & 68.00 \\*
CFM & 67.62 & 50.91 & 68.22 & 76.82 & 2.08 & 40.39 & 31.60 & 48.37 & 2.44 & 0.30 \\*
MLP & 86.00 & 84.21 & 86.00 & 30.68 & 3.73 & 61.85 & 47.18 & 65.88 & 4.34 & 0.30 \\*
MoT & 90.00 & \textbf{89.47} & 80.00 & 25.00 & 4.08 & 70.43 & 53.15 & 38.58 & 2.17 & 0.00 \\*
\method{} & \textbf{94.04} & 86.28 & \textbf{96.00} & \textbf{84.09} & \textbf{88.73} & \textbf{98.73} & \textbf{92.56} & \textbf{102.03} & \textbf{160.93} & \textbf{92.71} \\
\midrule
\multicolumn{11}{@{}l}{\textbf{Ministral 3 8B$\rightarrow$14B}} \\*
re-prefill & 77.20 & 65.61 & 78.14 & 77.19 & 87.34 & 76.62 & 89.23 & 73.25 & 28.67 & 68.00 \\*
CFM & 57.51 & 43.60 & 70.68 & 65.91 & 0.78 & 29.78 & 9.86 & 35.61 & 14.73 & 0.70 \\*
MLP & 82.00 & 92.11 & 86.00 & 28.41 & 8.51 & 83.17 & 62.15 & 79.82 & 14.95 & 0.63 \\*
MoT & 88.00 & 60.53 & 76.00 & 34.09 & 1.74 & 52.84 & 52.64 & 39.76 & 1.30 & 0.00 \\*
\method{} & \textbf{102.00} & \textbf{97.37} & \textbf{99.73} & \textbf{97.73} & \textbf{89.32} & \textbf{93.81} & \textbf{96.90} & \textbf{106.98} & \textbf{175.18} & \textbf{90.00} \\
\bottomrule
\end{tabular}
\endgroup
\end{table*}

\begin{table*}[!tbp]
\centering
\begingroup
\fontsize{8.5}{10.2}\selectfont
\setlength{\tabcolsep}{3pt}
\renewcommand{\arraystretch}{1.0}
\caption{Complete turn-indexed CoQA retention (\%). Columns give the handoff turn; each entry scores all post-handoff turns. Re-prefill is 100\%. Bold marks the highest transfer retention within each column and model pair.}\label{tab:accuracy-complete-coqa-turns}
\begin{tabular}{@{}p{0.170000\textwidth}>{\raggedleft\arraybackslash}p{\dimexpr0.083000\textwidth-2\tabcolsep\relax}>{\raggedleft\arraybackslash}p{\dimexpr0.083000\textwidth-2\tabcolsep\relax}>{\raggedleft\arraybackslash}p{\dimexpr0.083000\textwidth-2\tabcolsep\relax}>{\raggedleft\arraybackslash}p{\dimexpr0.083000\textwidth-2\tabcolsep\relax}>{\raggedleft\arraybackslash}p{\dimexpr0.083000\textwidth-2\tabcolsep\relax}>{\raggedleft\arraybackslash}p{\dimexpr0.083000\textwidth-2\tabcolsep\relax}>{\raggedleft\arraybackslash}p{\dimexpr0.083000\textwidth-2\tabcolsep\relax}>{\raggedleft\arraybackslash}p{\dimexpr0.083000\textwidth-2\tabcolsep\relax}>{\raggedleft\arraybackslash}p{\dimexpr0.083000\textwidth-2\tabcolsep\relax}>{\raggedleft\arraybackslash}p{\dimexpr0.083000\textwidth-2\tabcolsep\relax}@{}}
\toprule
Method / handoff turn & 1 & 2 & 3 & 4 & 5 & 6 & 7 & 8 & 9 & 10 \\
\midrule
\multicolumn{11}{@{}l}{\textbf{Qwen3 8B$\rightarrow$14B}} \\*
CFM & 25.96 & 45.12 & 50.80 & 54.14 & 45.67 & 49.11 & 53.90 & 40.49 & 36.72 & 42.37 \\*
MLP & 13.99 & 6.95 & 8.89 & 11.65 & 10.80 & 17.55 & 14.08 & 13.21 & 9.37 & 10.90 \\*
MoT & \textbf{79.45} & \textbf{76.02} & \textbf{67.22} & \textbf{58.43} & \textbf{55.89} & \textbf{65.73} & \textbf{56.92} & \textbf{50.83} & \textbf{49.59} & \textbf{50.91} \\*
\method{} & 27.05 & 39.37 & 31.61 & 33.04 & 36.30 & 36.66 & 32.01 & 26.99 & 22.77 & 24.90 \\
\midrule
\multicolumn{11}{@{}l}{\textbf{Qwen3 14B$\rightarrow$32B}} \\*
CFM & 103.07 & \textbf{107.25} & 102.11 & 111.18 & \textbf{115.10} & \textbf{119.22} & \textbf{120.63} & 101.40 & \textbf{120.61} & 105.16 \\*
MLP & \textbf{104.20} & 102.21 & \textbf{119.68} & \textbf{115.45} & 105.56 & 115.64 & 120.17 & \textbf{102.38} & 119.44 & \textbf{108.55} \\*
MoT & 100.76 & 96.17 & 111.82 & 99.64 & 95.99 & 98.26 & 97.78 & 92.82 & 112.57 & 96.76 \\*
\method{} & 92.60 & 90.75 & 99.56 & 91.95 & 91.21 & 108.20 & 117.43 & 91.05 & 105.62 & 100.27 \\
\midrule
\multicolumn{11}{@{}l}{\textbf{Llama 3.1 8B$\rightarrow$70B}} \\*
CFM & 69.56 & 95.02 & 66.71 & 79.21 & 72.79 & 73.91 & 71.84 & 70.27 & 58.52 & 81.55 \\*
MLP & 70.35 & \textbf{112.54} & 74.28 & 71.78 & 67.56 & 79.97 & 79.01 & 73.66 & 63.36 & 68.14 \\*
MoT & \textbf{96.14} & 100.05 & 93.49 & 97.33 & \textbf{97.32} & 91.74 & 95.34 & 91.68 & \textbf{96.73} & 89.12 \\*
\method{} & 88.07 & 92.58 & \textbf{93.82} & \textbf{98.73} & 97.16 & \textbf{99.70} & \textbf{106.90} & \textbf{105.70} & 92.27 & \textbf{100.26} \\
\midrule
\multicolumn{11}{@{}l}{\textbf{Ministral 3 3B$\rightarrow$14B}} \\*
CFM & 35.48 & 32.87 & 27.42 & 18.54 & 23.37 & 25.40 & 22.57 & 16.38 & 15.94 & 25.65 \\*
MLP & 59.63 & 69.50 & 64.57 & 58.47 & 48.59 & 59.37 & 59.43 & 50.79 & 52.29 & 52.67 \\*
MoT & 70.89 & \textbf{81.17} & 81.27 & 76.95 & 68.07 & 77.73 & 74.41 & 64.22 & 69.07 & 63.67 \\*
\method{} & \textbf{95.17} & 68.69 & \textbf{84.39} & \textbf{85.71} & \textbf{88.51} & \textbf{90.78} & \textbf{87.69} & \textbf{84.73} & \textbf{79.90} & \textbf{91.21} \\
\midrule
\multicolumn{11}{@{}l}{\textbf{Ministral 3 8B$\rightarrow$14B}} \\*
CFM & 10.60 & 11.34 & 8.19 & 7.55 & 8.34 & 9.63 & 9.17 & 4.71 & 6.70 & 10.06 \\*
MLP & 75.74 & 86.85 & 83.90 & 79.52 & 66.50 & 76.60 & 76.24 & 65.15 & 68.33 & 68.86 \\*
MoT & 79.03 & 91.59 & 70.64 & 71.89 & 70.08 & 72.53 & 70.70 & 66.70 & 69.26 & 58.80 \\*
\method{} & \textbf{100.71} & \textbf{107.08} & \textbf{96.77} & \textbf{99.33} & \textbf{92.05} & \textbf{94.98} & \textbf{94.23} & \textbf{85.78} & \textbf{90.13} & \textbf{94.85} \\
\bottomrule
\end{tabular}
\endgroup
\end{table*}

\newcommand{\CompleteAblationHeading}{%
\section{Complete Ablation Results}
\label{sec:appendix-system-ablation}

\paragraph{Value-map calibration.}
Table~\ref{tab:accuracy-complete-v-calibration} reports per-pair scores and
retention with and without value-map calibration on the ten single-turn
tasks.
\par
}

\begin{table*}[!tbp]
\CompleteAblationHeading
\centering
\begingroup
\fontsize{8.5}{10.2}\selectfont
\setlength{\tabcolsep}{2pt}
\renewcommand{\arraystretch}{1.0}
\caption{Value-map calibration ablation. Score rows report raw values (0--100), and retention rows report percentages. Retention without value-map calibration is normalized using re-prefill scores rounded to two decimal places. Bold marks the better retention of the two configurations, including ties.}\label{tab:accuracy-complete-v-calibration}
\begin{tabular}{@{}p{0.150000\textwidth}>{\raggedleft\arraybackslash}p{\dimexpr0.085000\textwidth-2\tabcolsep\relax}>{\raggedleft\arraybackslash}p{\dimexpr0.085000\textwidth-2\tabcolsep\relax}>{\raggedleft\arraybackslash}p{\dimexpr0.085000\textwidth-2\tabcolsep\relax}>{\raggedleft\arraybackslash}p{\dimexpr0.085000\textwidth-2\tabcolsep\relax}>{\raggedleft\arraybackslash}p{\dimexpr0.085000\textwidth-2\tabcolsep\relax}>{\raggedleft\arraybackslash}p{\dimexpr0.085000\textwidth-2\tabcolsep\relax}>{\raggedleft\arraybackslash}p{\dimexpr0.085000\textwidth-2\tabcolsep\relax}>{\raggedleft\arraybackslash}p{\dimexpr0.085000\textwidth-2\tabcolsep\relax}>{\raggedleft\arraybackslash}p{\dimexpr0.085000\textwidth-2\tabcolsep\relax}>{\raggedleft\arraybackslash}p{\dimexpr0.085000\textwidth-2\tabcolsep\relax}@{}}
\toprule
Configuration & \shortstack{Hella\\Swag} & ARC-C & \shortstack{Wino\\Grande} & MMLU & GSM8K & \shortstack{MMLU-\\Redux} & \shortstack{SQuAD\\v1.1} & QuALITY & Qasper & \shortstack{Hotpot\\QA-E} \\
\midrule
\multicolumn{11}{@{}l}{\textbf{Qwen3 8B$\rightarrow$14B}} \\*
re-prefill (score) & 78.12 & 61.18 & 74.98 & 83.33 & 96.89 & 68.02 & 21.78 & 80.49 & 39.35 & 66.06 \\*
w/o V cal. (score) & 59.38 & 46.88 & 53.13 & 81.25 & 6.25 & 59.38 & 6.89 & 50.00 & 11.34 & 14.28 \\*
w/o V cal. (\%) & 76.01 & 76.63 & 70.86 & \textbf{97.50} & 6.45 & 87.30 & 31.63 & 62.12 & 28.82 & 21.62 \\*
Full (\%) & \textbf{96.00} & \textbf{93.46} & \textbf{93.75} & 95.79 & \textbf{100.00} & \textbf{94.59} & \textbf{85.39} & \textbf{97.09} & \textbf{96.78} & \textbf{88.02} \\
\midrule
\multicolumn{11}{@{}l}{\textbf{Qwen3 14B$\rightarrow$32B}} \\*
re-prefill (score) & 78.12 & 54.69 & 65.59 & 85.26 & 45.26 & 66.65 & 26.20 & 84.37 & 41.46 & 62.25 \\*
w/o V cal. (score) & 71.88 & 59.38 & 68.75 & 81.25 & 56.25 & 75.00 & 23.29 & 65.63 & 27.86 & 42.04 \\*
w/o V cal. (\%) & 92.01 & 108.58 & 104.82 & 95.30 & 124.28 & \textbf{112.53} & 88.89 & 77.79 & 67.20 & 67.53 \\*
Full (\%) & \textbf{104.00} & \textbf{111.43} & \textbf{111.90} & \textbf{97.54} & \textbf{131.99} & 101.04 & \textbf{97.36} & \textbf{103.70} & \textbf{93.34} & \textbf{96.64} \\
\midrule
\multicolumn{11}{@{}l}{\textbf{Llama 3.1 8B$\rightarrow$70B}} \\*
re-prefill (score) & 81.25 & 60.92 & 81.22 & 83.33 & 80.14 & 77.22 & 82.87 & 83.70 & 16.20 & 15.77 \\*
w/o V cal. (score) & 68.75 & 37.50 & 50.00 & 37.50 & 6.25 & 28.13 & 13.08 & 37.50 & 2.05 & 0.45 \\*
w/o V cal. (\%) & 84.62 & 61.56 & 61.56 & 45.00 & 7.80 & 36.43 & 15.78 & 44.80 & 12.65 & 2.85 \\*
Full (\%) & \textbf{100.00} & \textbf{100.00} & \textbf{100.00} & \textbf{100.00} & \textbf{93.60} & \textbf{85.42} & \textbf{83.16} & \textbf{96.49} & \textbf{109.75} & \textbf{93.85} \\
\midrule
\multicolumn{11}{@{}l}{\textbf{Ministral 3 3B$\rightarrow$14B}} \\*
re-prefill (score) & 77.20 & 65.61 & 78.14 & 77.19 & 87.34 & 76.62 & 89.23 & 73.25 & 28.67 & 68.00 \\*
w/o V cal. (score) & 68.75 & 43.75 & 78.13 & 68.75 & 65.63 & 75.00 & 74.67 & 71.88 & 22.39 & 37.52 \\*
w/o V cal. (\%) & 89.05 & 66.68 & \textbf{99.99} & \textbf{89.07} & 75.14 & 97.89 & 83.68 & 98.13 & 78.10 & 55.18 \\*
Full (\%) & \textbf{94.04} & \textbf{86.28} & 96.00 & 84.09 & \textbf{88.73} & \textbf{98.73} & \textbf{92.56} & \textbf{102.03} & \textbf{160.93} & \textbf{92.71} \\
\midrule
\multicolumn{11}{@{}l}{\textbf{Ministral 3 8B$\rightarrow$14B}} \\*
re-prefill (score) & 77.20 & 65.61 & 78.14 & 77.19 & 87.34 & 76.62 & 89.23 & 73.25 & 28.67 & 68.00 \\*
w/o V cal. (score) & 75.00 & 56.25 & 81.25 & 75.00 & 71.88 & 68.75 & 83.44 & 65.63 & 28.37 & 48.44 \\*
w/o V cal. (\%) & 97.15 & 85.73 & \textbf{103.98} & 97.16 & 82.30 & 89.73 & 93.51 & 89.60 & 98.95 & 71.24 \\*
Full (\%) & \textbf{102.00} & \textbf{97.37} & 99.73 & \textbf{97.73} & \textbf{89.32} & \textbf{93.81} & \textbf{96.90} & \textbf{106.98} & \textbf{175.18} & \textbf{90.00} \\
\bottomrule
\end{tabular}
\endgroup
\end{table*}

\begin{table*}[!tbp]
\begin{minipage}[t]{\columnwidth}
\vspace{0pt}
\paragraph{Execution components.}
The following tables give detailed results for pipeline overlap, path
scheduling and request concurrency on Qwen3 14B$\rightarrow$32B.
Direct installation and dynamic buffer allocation are reported in
Tables~\ref{tab:direct-installation} and~\ref{tab:dynamic-buffer-allocation}.
Full denotes complete \method{} within each matched experiment.
Each table retains its own workload and configuration, with $C$ denoting
request concurrency; requests arrive serially in groups of $C$. All TTFT
values are in milliseconds.

\paragraph{Pipeline overlap.}
Serial execution disables overlap among reconstruction, transfer and
installation within each request.
\par
\centering
\begingroup
\fontsize{8.5}{10.2}\selectfont
\setlength{\tabcolsep}{3pt}
\renewcommand{\arraystretch}{1.0}
\caption{Pipeline overlap: TTFT (ms), 512 requests per configuration.}
\label{tab:system-ablation-pipeline-overlap}
\begin{tabular*}{\columnwidth}{@{\extracolsep{\fill}}lrrrrr@{}}
\toprule
 & & \multicolumn{2}{c}{Full} & \multicolumn{2}{c}{Serial} \\
\cmidrule(lr){3-4}\cmidrule(l){5-6}
History & $C$ & Mean & P99 & Mean & P99 \\
\midrule
8K & 1 & 219.00 & 364.00 & 250.00 & 344.00 \\
 & 2 & 254.00 & 392.00 & 296.00 & 485.00 \\
 & 4 & 329.00 & 711.00 & 379.00 & 780.00 \\
 & 8 & 445.00 & 1,323.00 & 531.00 & 1,249.00 \\
 & 16 & 1,663.00 & 3,361.00 & 2,207.00 & 3,944.00 \\
\midrule
32K & 1 & 395.00 & 589.00 & 543.00 & 653.00 \\
 & 2 & 515.00 & 776.00 & 656.00 & 1,056.00 \\
 & 4 & 896.00 & 2,250.00 & 1,380.00 & 2,894.00 \\
 & 8 & 2,693.00 & 4,461.00 & 3,897.00 & 4,865.00 \\
 & 16 & 7,567.00 & 10,108.00 & 9,189.00 & 10,678.00 \\
\bottomrule
\end{tabular*}
\endgroup

\end{minipage}\hfill
\begin{minipage}[t]{\columnwidth}
\vspace{0pt}
\paragraph{Path scheduling.}
Fixed transfer disables path scheduling while retaining memory admission
and re-prefill fallback.
Each batch equally mixes 512-token, 1K, 2K, 4K and 8K histories.
\par
\centering
\begingroup
\fontsize{8.5}{10.2}\selectfont
\setlength{\tabcolsep}{3pt}
\renewcommand{\arraystretch}{1.0}
\caption{Path scheduling: TTFT (ms), 512 requests per configuration.}
\label{tab:system-ablation-path-scheduling}
\begin{tabular*}{\columnwidth}{@{\extracolsep{\fill}}rrrrr@{}}
\toprule
 & \multicolumn{2}{c}{Full} & \multicolumn{2}{c}{Fixed transfer} \\
\cmidrule(lr){2-3}\cmidrule(l){4-5}
$C$ & Mean & P99 & Mean & P99 \\
\midrule
4 & 247.00 & 429.20 & 662.10 & 1,457.70 \\
8 & 2,493.00 & 8,219.00 & 4,675.00 & 15,514.00 \\
16 & 16,411.00 & 33,617.00 & 20,613.00 & 43,451.00 \\
\bottomrule
\end{tabular*}
\endgroup

\paragraph{Request concurrency.}
The ablation admits only one transfer pipeline while preserving the offered
request load.
\par
\centering
\begingroup
\fontsize{8.5}{10.2}\selectfont
\setlength{\tabcolsep}{3pt}
\renewcommand{\arraystretch}{1.0}
\caption{Request concurrency: TTFT (ms).}
\label{tab:system-ablation-request-concurrency}
\begin{tabular*}{\columnwidth}{@{\extracolsep{\fill}}lrrrrr@{}}
\toprule
 & & \multicolumn{2}{c}{Full} & \multicolumn{2}{c}{Single pipeline} \\
\cmidrule(lr){3-4}\cmidrule(l){5-6}
History & $C$ & Mean & P99 & Mean & P99 \\
\midrule
2K & 4 & 165.38 & 400.50 & 323.79 & 547.52 \\
 & 8 & 307.72 & 658.01 & 547.30 & 928.80 \\
 & 16 & 617.80 & 1,049.00 & 846.20 & 1,533.90 \\
\midrule
8K & 4 & 267.09 & 527.68 & 436.85 & 706.36 \\
 & 8 & 479.85 & 835.91 & 730.27 & 1,325.14 \\
 & 16 & 918.90 & 1,271.70 & 1,714.70 & 2,643.90 \\
\midrule
32K & 4 & 685.86 & 1,286.18 & 900.36 & 1,470.32 \\
 & 8 & 1,088.55 & 1,691.97 & 1,578.82 & 2,791.80 \\
 & 16 & 2,170.10 & 4,596.00 & 3,236.20 & 6,165.10 \\
\bottomrule
\end{tabular*}
\endgroup

\end{minipage}
\end{table*}

\begin{table*}[!tbp]
\begin{minipage}[t]{\columnwidth}
\vspace{0pt}
\paragraph{Compact mapper.}
We replace only the mapper on the optimized handoff path. CFM's required
source-side KV gathering and TP synchronization are included in handoff
TTFT. Both configurations use identical calibration data, dtypes, buffer
budgets and request traces, while retaining their respective fitting and
calibration procedures. Table~\ref{tab:cfm-mapper-optimized-path} reports
all mean and P99 TTFT ratios.
\par
\centering
\begingroup
\fontsize{8.5}{10.2}\selectfont
\setlength{\tabcolsep}{4pt}
\caption{Compact mapping lowers mean and P99 TTFT with the same system optimizations. Ratios are \method{}/(CFM mapper + \method{} path), with $C$ denoting concurrency.}
\label{tab:cfm-mapper-optimized-path}
\begin{tabular}{@{}lrrrrr@{}}
\toprule
History & $C=1$ & $C=2$ & $C=4$ & $C=8$ & $C=16$ \\
\midrule
\multicolumn{6}{@{}l}{\textit{Mean}} \\
512 & 0.773 & 0.933 & 0.967 & 0.955 & 0.944 \\
1K & 0.809 & 0.908 & 0.887 & 0.830 & 0.826 \\
2K & 0.700 & 0.824 & 0.813 & 0.765 & 0.695 \\
4K & 0.647 & 0.643 & 0.684 & 0.729 & 0.724 \\
8K & 0.529 & 0.513 & 0.559 & 0.599 & 0.626 \\
16K & 0.450 & 0.425 & 0.462 & 0.501 & 0.515 \\
32K & 0.379 & 0.352 & 0.387 & 0.426 & 0.457 \\
\midrule
\multicolumn{6}{@{}l}{\textit{P99}} \\
512 & 0.789 & 0.792 & 0.992 & 0.706 & 0.872 \\
1K & 0.540 & 0.966 & 0.904 & 0.688 & 0.693 \\
2K & 0.521 & 0.671 & 0.752 & 0.810 & 0.866 \\
4K & 0.452 & 0.438 & 0.620 & 0.993 & 0.754 \\
8K & 0.459 & 0.543 & 0.712 & 0.671 & 0.635 \\
16K & 0.391 & 0.702 & 0.555 & 0.534 & 0.535 \\
32K & 0.346 & 0.439 & 0.453 & 0.466 & 0.492 \\
\bottomrule
\end{tabular}
\endgroup

\end{minipage}\hfill
\begin{minipage}[t]{\columnwidth}
\vspace{0pt}
\paragraph{Block size.}
Table~\ref{tab:system-ablation-block-size} varies reconstruction blocks
from 512 to 4,096 tokens. Larger blocks reduce handoff latency but require
proportionally more transfer-buffer memory, leaving less memory for
inference. Doubling the adopted block from 2,048 to 4,096 tokens doubles
buffer memory for 1.9\% and 12.0\% lower TTFT at 8K and 32K, respectively.
The 2,048-token block thus balances latency against memory, staying within
13.6\% of the fastest measured TTFT at half the buffer memory.
\par
\centering
\begingroup
\fontsize{8.5}{10.2}\selectfont
\setlength{\tabcolsep}{3pt}
\renewcommand{\arraystretch}{1.0}
\caption{Reconstruction block size: median TTFT and relative transfer-buffer
memory, normalized to 2,048-token blocks. At fixed mapper and transfer
concurrency, buffer capacity scales linearly with block size. The adopted block is bold.}
\label{tab:system-ablation-block-size}
\begin{tabular*}{\columnwidth}{@{\extracolsep{\fill}}lrrr@{}}
\toprule
 & \multicolumn{2}{c}{Median TTFT (ms)} & Buffer memory \\
\cmidrule(lr){2-3}
Block (tokens) & 8K & 32K & (relative) \\
\midrule
512 & 188.89 & 563.31 & 0.25 \\
1,024 & 140.84 & 368.93 & 0.50 \\
\textbf{2,048} & \textbf{123.71} & \textbf{291.97} & \textbf{1.00} \\
4,096 & 121.34 & 257.05 & 2.00 \\
\bottomrule
\end{tabular*}
\endgroup

\end{minipage}
\end{table*}

\FloatBarrier
\begin{table*}[!tbp]
\begin{minipage}{\textwidth}
\section{Distributed Handoffs Across Two Servers}
\label{sec:appendix-multinode}

Section~\ref{sec:system-implementation} describes the cross-node transport,
and Section~\ref{sec:eval-multinode} reports the measurement protocol and
heatmap. Table~\ref{tab:multinode-tp8-ttft} gives all mean and P99 TTFT
ratios for the two-server TP8 experiment.
\end{minipage}
\par\medskip
\par
\centering
\begingroup
\fontsize{8.5}{10.2}\selectfont
\setlength{\tabcolsep}{5pt}
\renewcommand{\arraystretch}{1.0}
\caption{Two-server TP8 TTFT ratios (\method{}/re-prefill, lower is better). Each cell gives mean/P99, and $C$ denotes request concurrency.}
\label{tab:multinode-tp8-ttft}
\makebox[\linewidth][c]{%
\begin{tabular}{@{}lrrrrr@{}}
\toprule
History & $C=1$ & $C=2$ & $C=4$ & $C=8$ & $C=16$ \\
\midrule
512 & 2.741/2.256 & 2.584/2.278 & 2.566/2.672 & 2.679/4.075 & 2.455/3.434 \\
1K & 1.805/1.709 & 1.717/1.482 & 1.584/2.483 & 1.522/1.782 & 1.492/1.995 \\
2K & 1.265/1.295 & 0.993/1.008 & 0.852/1.003 & 0.846/1.096 & 0.893/1.145 \\
4K & 0.824/0.717 & 0.529/0.908 & 0.525/0.633 & 0.579/0.974 & 0.607/0.684 \\
8K & 0.478/0.795 & 0.313/0.389 & 0.314/0.391 & 0.333/0.374 & 0.343/0.376 \\
16K & 0.342/0.457 & 0.289/0.361 & 0.259/0.254 & 0.226/0.218 & 0.212/0.210 \\
32K & 0.237/0.259 & 0.219/0.252 & 0.194/0.193 & 0.177/0.177 & 0.167/0.169 \\
\bottomrule
\end{tabular}}
\endgroup

\end{table*}

\end{document}